\documentclass[letterpaper]{article}
\PassOptionsToPackage{table}{xcolor}
\usepackage[preprint]{arxiv}
\usepackage[hyphens]{url}
\usepackage{graphicx}
\usepackage{natbib}
\usepackage{caption}
\usepackage{booktabs}
\usepackage{amsmath,amssymb,bm}
\usepackage{multirow}
\usepackage{makecell}
\usepackage[table]{xcolor}
\usepackage{threeparttable}
\usepackage{array}
\usepackage{tabularx}
\newcommand{\PaperTitle}{Act-HiMotif: Hierarchical Daily Activity Schedule Generation via Learned Activity-Structure Motifs}
\title{\PaperTitle}
\author{
    Yushi Bai\textsuperscript{\rm 1,2},
    Kang Liu\textsuperscript{\rm 1,3,4}\thanks{Corresponding author: kang.liu@siat.ac.cn.},
    Ling Yin\textsuperscript{\rm 1,3,4},
    Wei Wang\textsuperscript{\rm 5},
    Yilin Lyu\textsuperscript{\rm 1,3,6}
}
\affiliations{
    \textsuperscript{\rm 1}Shenzhen Institutes of Advanced Technology, Chinese Academy of Sciences, Shenzhen, China\\
    \textsuperscript{\rm 2}School of Computer Science and Engineering, Macau University of Science and Technology, Macau, China\\
    \textsuperscript{\rm 3}Shenzhen University of Advanced Technology, Shenzhen, China\\
    \textsuperscript{\rm 4}University of Chinese Academy of Sciences, Beijing, China\\
    \textsuperscript{\rm 5}City University of Hong Kong, Hong Kong, China\\
    \textsuperscript{\rm 6}Beijing Normal-Hong Kong Baptist University, Zhuhai, China\\
    3250009371@student.must.edu.mo, kang.liu@siat.ac.cn, yinling@siat.ac.cn\\
    wwang326@cityu.edu.hk, lyuyilin@uic.edu.cn
}
\newcommand{\PaperAuthors}{Yushi Bai, Kang Liu, Ling Yin, Wei Wang, Yilin Lyu}

\begin{document}
\maketitle

\begin{abstract}
Generating daily activity schedules from individual attributes (e.g., age and income) is valuable for activity-based travel modeling and human behavior simulation. The task is challenging because exact activity structures follow a highly imbalanced, long-tailed distribution, while the same structure can admit multiple duration allocations under a fixed full-day budget. Direct generation must therefore learn many exact activity structures from highly uneven observations while modeling their possible temporal realizations. However, many exact structures share recurring activity organization. Based on this observation, we propose Act-HiMotif, a hierarchical motif-based generative model that introduces learned, reusable activity-structure motifs as an intermediate representation between individual profiles and specific activity structures. Act-HiMotif first distinguishes home-only days from active days, then samples a motif from a profile-conditioned prior to guide activity-structure generation, and finally applies a budget-constrained duration allocator to generate activity durations that sum exactly to the full-day horizon. This hierarchy models home-only routing, activity-structure generation, and duration allocation at their corresponding levels, while motifs allow related structures to share statistical support without losing their distinct sequences. Experiments on large-scale travel survey data show that Act-HiMotif outperforms strong baselines in both in-distribution and spatial generalization settings, demonstrating the effectiveness of hierarchical generation and reusable activity-structure motifs for producing realistic profile-conditioned daily activity schedules.
\end{abstract}

\section{Introduction}
Daily activity schedules describe which activities individuals perform, in what order, and for how long \citep{Krishna2018ActivityDuration, Yuan2024NeedDynamics, Yuan2022ActivityTrajectory}. They provide important behavioral inputs for activity-based travel demand modeling, synthetic population simulation, and human mobility analysis \citep{Bowman2001ActivityBased, Luca2021MobilitySurvey, Kapp2024SyntheticMobilitySurvey, Fan2026UrbanMixing}. Because household travel surveys are costly to collect at scale \citep{BaduMarfo2020DPDiary, Lyu2026FLAIR}, profile-conditioned daily activity schedule generation seeks to synthesize plausible complete schedules from attributes such as age, employment status, and income \citep{Shone2025ActVAE, Liao2026DeepActivity, Li2024MobAgent}, enabling activity behavior to be assigned to synthetic populations and profile-aware agents \citep{Drchal2019DataDrivenScheduler, Shone2025DGML, Liu2024Act2Loc, Park2023GenerativeAgents, Wang2024LLMob}.

\begin{figure}[!t]
    \centering
    \includegraphics[width=\linewidth]{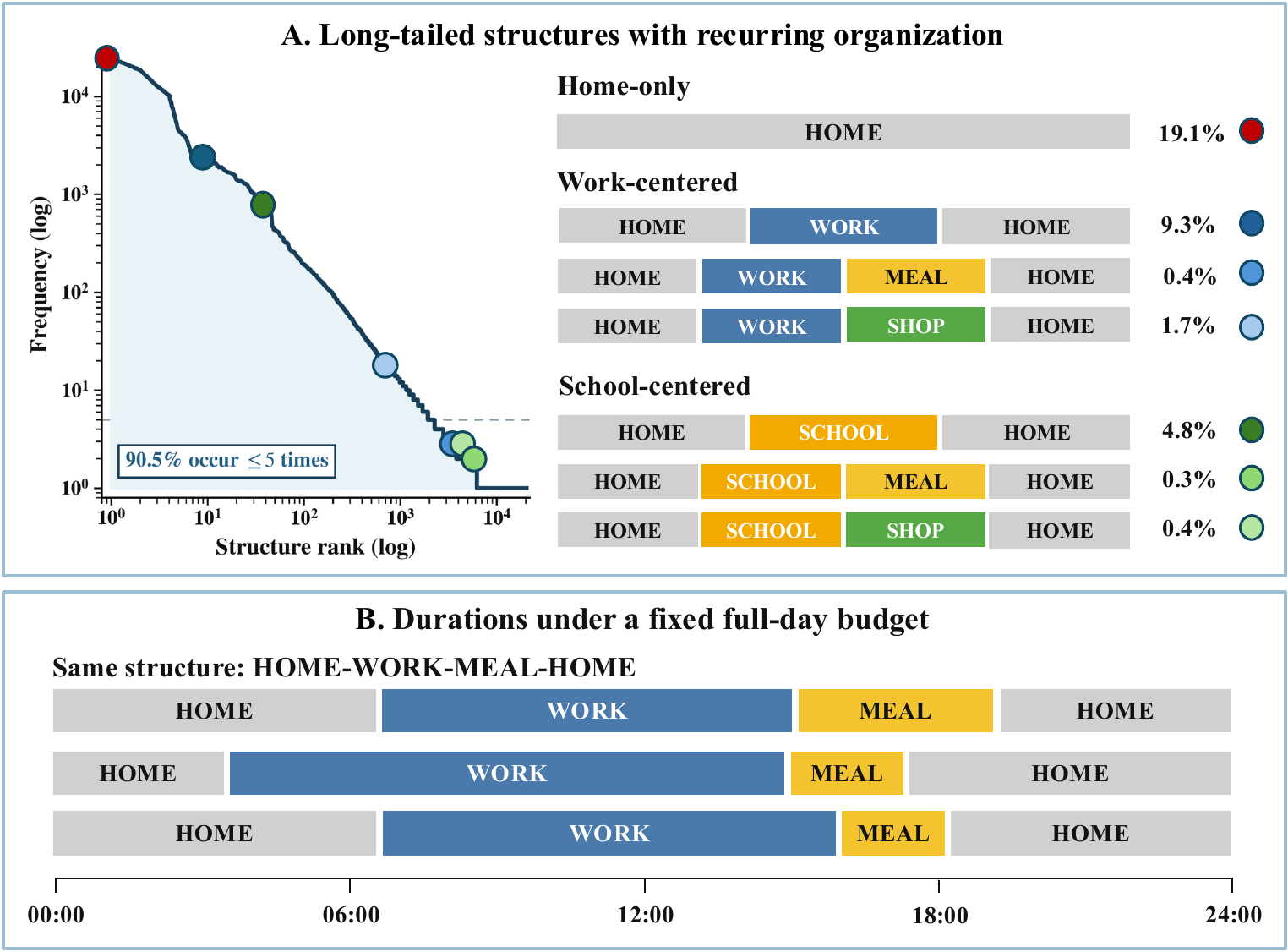}
    \caption{Motivating properties of daily activity schedules: A. long-tailed activity structures with recurring patterns and B. multiple duration allocations for the same structure under the full-day time constraint.}
    \label{fig1}
\end{figure}

Traditional activity-based models assemble schedules through prespecified behavioral components and context-specific calibration, while recent deep generative methods learn schedules directly from travel records but generally model complete schedules or activity chains as a single generation target \citep{Pougala2023OASIS, Bowman2001ActivityBased, Arentze2004ALBATROSS, Bhat2008MDCEV, Shone2025ActVAE, Liao2026DeepActivity}. Prior mobility studies have also identified recurring daily motifs and long-tailed visitation patterns \citep{Schneider2013MobilityMotifs, Xu2024LongTailMobility}. Figure~\ref{fig1} highlights two difficulties of activity-schedule generation. 
First, exact activity structures are highly imbalanced and long-tailed: in the 2017 U.S. National Household Travel Survey (NHTS), the home-only structure accounts for 19.1\% of person-days, whereas 90.5\% of distinct structures occur no more than five times. Nevertheless, many structures share recurring work- or school-centered organization. Second, the same ordered activity structure can have multiple duration allocations, all of which must satisfy the fixed full-day budget. Consequently, direct generation must learn diverse activity structures from highly uneven observations while simultaneously producing temporally valid schedules.

To address these challenges, we propose Act-HiMotif based on two key ideas. The first is a hierarchical formulation tailored to the distinct components of a daily schedule. The model must determine whether a day is home-only or active, generate the ordered activities for an active day, and allocate the full-day duration among them. Treating these decisions as a flat generation target requires a single model to simultaneously handle the prominent home-only outcome, combinatorial activity structures, and the global temporal constraint. A coarse-to-fine formulation instead models these sources of variation at their corresponding levels while producing a complete and valid schedule. The second key idea is the use of activity-structure motifs. Although exact activity structures follow a long-tailed distribution, many rare structures are variations of recurring activity organizations. For example, the home--work--home, home--work--meal--home, and home--work--shop--home structures in Figure~\ref{fig1}A share a recognizable work-centered organization. Modeling each exact structure independently prevents rare structures from benefiting from related observations, whereas assigning them to a single predefined category would remove meaningful differences in activity presence and ordering. Learned activity-structure motifs allow related structures to share statistical support while preserving their distinct activity sequences.

Act-HiMotif integrates these ideas into a hierarchical motif-based generation process. A Profile Encoder represents the individual profile and day-of-week information, and an Active-Day Gate first distinguishes home-only from active days. A home-only outcome directly produces the full-day home schedule. For an active day, a Profile-Conditioned Motif Prior samples a motif from the discrete space learned from duration-free activity structures by the Vector-Quantized Motif Learner. The Motif-Conditioned Structure Decoder then generates an exact activity structure, after which the Budget-Constrained Duration Allocator assigns positive episode durations that sum exactly to the full-day horizon. The resulting hierarchy separates home-only routing, activity-structure generation, and constrained duration allocation, while motif conditioning enables statistical sharing across related structures.

Experiments on 264,234 person-day schedules derived from the 2017 U.S. NHTS show that Act-HiMotif consistently improves the structural and temporal fidelity of profile-conditioned daily activity schedules under both in-distribution and spatial generalization evaluation.

Our main contributions are summarized as follows:
\begin{itemize}
    \item We develop a hierarchical profile-conditioned generative framework that separates home-only routing, exact activity-structure generation, and budget-constrained duration allocation, producing complete schedules under a fixed full-day budget.
    
    \item We introduce learned, reusable activity-structure motifs as an intermediate representation between profile conditions and exact activity structures, enabling information sharing across long-tailed structures while preserving their distinct activity sequences.

    \item We conduct extensive experiments on large-scale travel survey data and demonstrate consistent improvements over strong baselines under both in-distribution and spatial generalization settings, with code provided in the Supplementary Material.
\end{itemize}

\section{Related Work}

\paragraph{Traditional activity-based schedule generation.}
Traditional activity-based travel demand models generate daily schedules by decomposing activity participation and scheduling into linked behavioral decisions concerning activities, tours, destinations, modes, and timing \citep{Bowman2001ActivityBased}. Computational process models such as ALBATROSS implement these decisions through rule- or decision-tree-based mechanisms, while MDCEV formulations jointly model participation in multiple activities and the allocation of finite temporal resources \citep{Arentze2004ALBATROSS, Bhat2008MDCEV}. These approaches provide behaviorally interpretable scheduling mechanisms and explicitly represent behavioral or temporal constraints. However, they rely on prespecified decision structures, behavioral assumptions, and substantial context-specific calibration.

\paragraph{Deep generative activity-schedule generation.}
Deep generative methods learn activity participation, ordering, and timing directly from observed diaries, reducing the need for manually specified behavioral rules. Early data-driven schedulers use recurrent models to generate sequential activity decisions from travel-survey records \citep{BaduMarfo2020DPDiary, Drchal2019DataDrivenScheduler, Shone2025DGML}. More recent methods introduce stochastic latent representations or structured behavioral mechanisms: ActVAE employs a conditional variational autoencoder to generate diverse schedules from individual attributes, NeedDynamics models activity generation through evolving human needs, and DeepActivity generates profile-conditioned activity and location chains \citep{Shone2025ActVAE, Yuan2024NeedDynamics, Liao2026DeepActivity}. Although these approaches improve data-driven activity generation, they generally model complete schedules or activity chains directly. 

\paragraph{LLM-based activity and travel-diary generation.}
Recent LLM-based approaches generate daily activities and travel diaries by incorporating semantic knowledge, individual profiles, and explicit behavioral reasoning. LLMob represents individuals as language-model agents whose mobility decisions are guided by habitual patterns and current motivations \citep{Wang2024LLMob}. MobAgent uses individual profiles to extract mobility patterns and generate personalized travel diaries, while TravelReasoner reformulates trip-chain generation as an explicit reasoning process \citep{Li2024MobAgent, Liu2026TravelReasoner}. ActivityEditor further employs generation and editing agents to iteratively refine activity chains under behavioral and physical constraints \citep{Yang2026ActivityEditor}. These methods enhance semantic and intention-aware activity generation, but typically depend on language-model reasoning or iterative refinement.

\section{Problem Formulation}
\label{sec:problem_formulation}

We study profile-conditioned daily activity schedule generation. Each person-day is associated with a profile condition $\mathbf{x}=(x_1,\ldots,x_F)\in\mathcal{X}$, which includes individual attributes and day-of-week information. Here, $F$ is the number of conditioning fields, $x_f$ is the value of the $f$-th field, and $\mathcal{X}$ is the profile space.

A daily activity schedule contains a variable number of ordered activity episodes and is represented as
\begin{equation}
Y=\big((c_1,d_1),\ldots,(c_K,d_K)\big)=(S,\mathbf{d}),
\label{eq:schedule_representation}
\end{equation}
where $K$ is the number of episodes, $c_k\in\mathcal{C}$ is the activity type of episode $k$, and $d_k\in\mathbb{N}_{+}$ is its duration in discrete time slots. The activity structure $S=(c_1,\ldots,c_K)$ records the ordered activity types, and the duration vector $\mathbf{d}=(d_1,\ldots,d_K)$ specifies their temporal allocation. For a full-day horizon of $T$ slots, a valid schedule satisfies
\begin{equation}
 c_k\neq c_{k+1}\;(1\leq k<K),\qquad d_k\geq 1,\qquad \sum_{k=1}^{K}d_k=T.
\label{eq:schedule_validity}
\end{equation}
The first condition reflects the canonical episode-level representation obtained by merging adjacent occurrences of the same activity type, while the remaining conditions ensure that all episodes have positive duration and exactly cover the full-day horizon.

Given an observed dataset $\mathcal{D}=\{(\mathbf{x}^{(i)},Y^{(i)})\}_{i=1}^{N}$, our goal is to learn a conditional distribution $P_{\theta}(Y\mid\mathbf{x})$ over valid daily schedules, where $\theta$ denotes the learnable model parameters. This distributional formulation allows the same or similar profile conditions to correspond to multiple plausible activity structures and duration allocations.

\section{Methodology}
\label{sec:methodology}

\begin{figure*}[htbp]
    \centering
    \includegraphics[width=\linewidth]{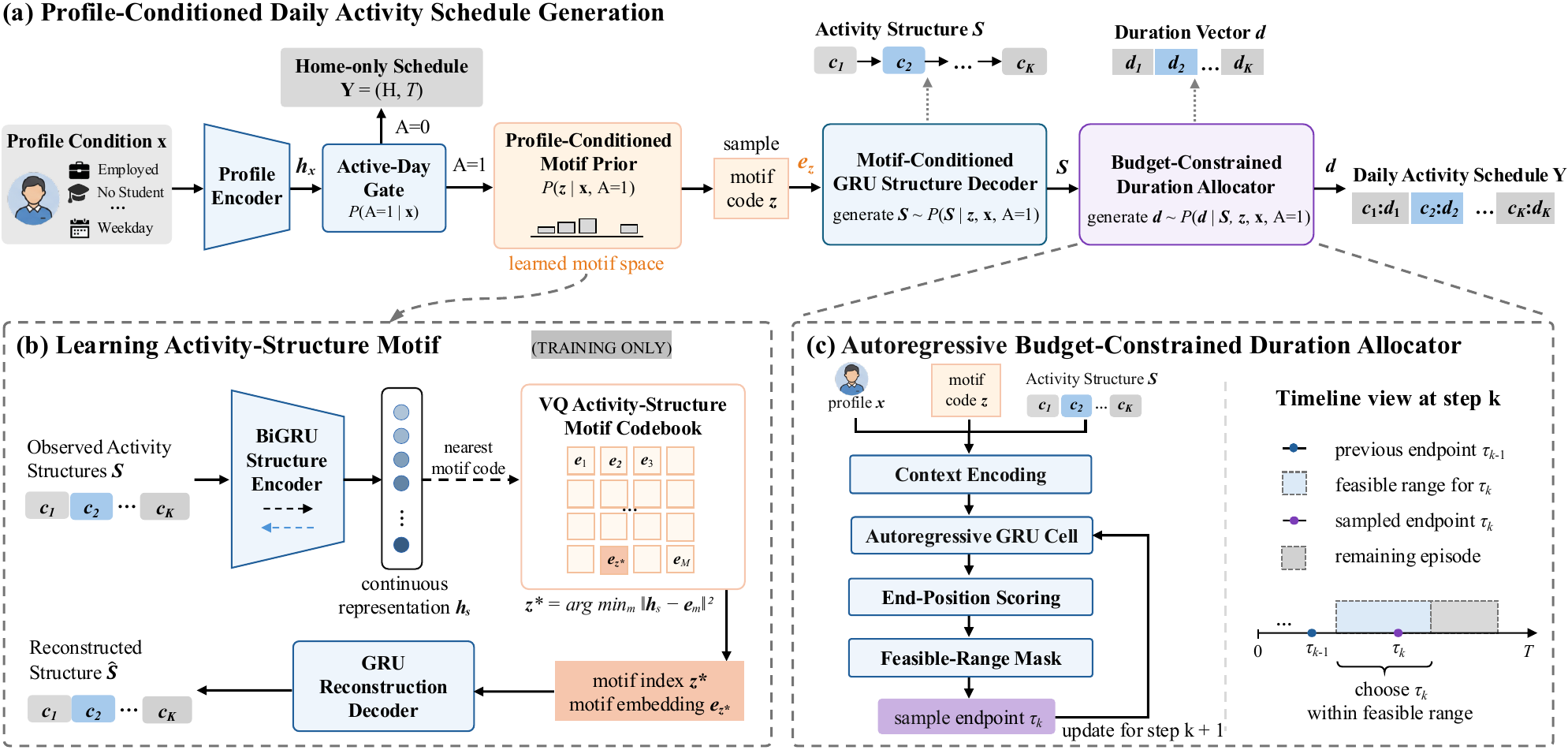}
    \caption{Overview of Act-HiMotif: (a) hierarchical profile-conditioned schedule generation, (b) training-only learning of reusable activity-structure motifs, and (c) autoregressive budget-constrained duration allocation.}
    \label{fig2}
\end{figure*}



Act-HiMotif generates a daily activity schedule in a coarse-to-fine manner. Given a profile condition, it first distinguishes home-only from active days. For an active day, it samples a reusable activity-structure motif, generates an activity structure conditioned on the profile and motif, and allocates episode durations under the fixed full-day budget. Figure~\ref{fig2}(a) presents the complete generation path, while Figures~\ref{fig2}(b) and~\ref{fig2}(c) detail the training-only motif learner and the budget-constrained duration allocator, respectively.

\subsection{Profile-Conditioned Activity-Structure Generation}

\subsubsection{Profile Encoding and Active-Day Gate}
\label{sec:profile_active_gate}

Each categorical profile field, including day-of-week information, is mapped through a field-specific embedding. The field embeddings are concatenated and transformed by a multilayer perceptron $f_x$ to obtain the shared profile representation $\mathbf h_x=f_x(\mathbf x)$, which conditions the subsequent generation modules.

Let $A\in\{0,1\}$ be the active-day indicator, where $A=0$ denotes a home-only day and $A=1$ denotes a day containing at least one non-home activity. The Active-Day Gate computes
\begin{equation}
 p_A(\mathbf x)
 =P_{\theta_A}(A=1\mid\mathbf x)
 =\sigma\!\left(g_A(\mathbf h_x)\right),
\label{eq:active_day_gate}
\end{equation}
where $g_A$ is a binary prediction head, $\sigma(\cdot)$ is the sigmoid function, and $p_A(\mathbf x)$ is the active-day probability. 
If $A=0$, the model directly returns the full-day home schedule $Y=((\mathrm H,T))$. Conditional on $A=1$, the remaining generation process factorizes as
\begin{equation}
\begin{aligned}
&P_{\theta}(z,S,\mathbf d\mid\mathbf x,A=1) = P_{\theta_z}(z\mid\mathbf x,A=1) \\
&\cdot P_{\theta_S}(S\mid z,\mathbf x,A=1)\cdot P_{\theta_D}(\mathbf d\mid S,z,\mathbf x,A=1),
\end{aligned}
\label{eq:hierarchical_factorization}
\end{equation}
where the motif $z$ captures reusable structural organization, $S$ specifies its exact activity sequence, and $\mathbf d$ provides a valid temporal realization under the full-day budget.

\subsubsection{Learning Reusable Activity-Structure Motifs}
\label{sec:motif_learning}

We learn motifs from duration-free active-day structures so that they capture recurring activity presence and ordering. 
Given an observed active-day structure $S=(c_1,\ldots,c_K)$, a bidirectional GRU encodes the activity-token sequence into a continuous structure representation $\mathbf h_S$. Let $\{\mathbf e_m\}_{m=1}^{M}$ be a codebook of $M$ learnable motif embeddings. The structure is assigned to its nearest motif according to
\begin{equation}
 z^{\star}
 =\arg\min_{1\leq m\leq M}
 \left\|\mathbf h_S-\mathbf e_m\right\|_2^2,
\label{eq:motif_quantization}
\end{equation}
where $z^{\star}$ is the nearest-motif index and $\mathbf e_{z^{\star}}$ is the selected motif embedding.

A GRU reconstruction decoder is initialized from $\mathbf e_{z^{\star}}$ and reconstructs the exact activity structure autoregressively. Let $c_0=\langle\mathrm{BOS}\rangle$ and $c_{K+1}=\langle\mathrm{EOS}\rangle$. The token-level reconstruction loss is
\begin{equation}
 \mathcal L_{\mathrm{rec}}
 =-\sum_{k=1}^{K+1}
 \log P_{\phi}\!\left(
 c_k\mid c_{0:k-1},\mathbf e_{z^{\star}}
 \right),
\label{eq:motif_reconstruction_loss}
\end{equation}
where $c_{0:k-1}$ denotes the observed tokens preceding position $k$. Reconstructing the exact sequence encourages each motif to retain information about activity ordering.

We use the standard nearest-neighbor vector-quantization bottleneck of VQ-VAE~\citep{vanDenOord2017VQVAE}, while adopting a task-specific BiGRU--GRU autoencoder for activity structures. Its core objective is
\begin{equation}
\begin{aligned}
\mathcal L_{\mathrm{VQ}}
={}&\mathcal L_{\mathrm{rec}}
+\left\|\operatorname{sg}[\mathbf h_S]-\mathbf e_{z^{\star}}\right\|_2^2
+\beta\left\|\mathbf h_S-\operatorname{sg}[\mathbf e_{z^{\star}}]\right\|_2^2,
\end{aligned}
\label{eq:vq_core_objective}
\end{equation}
where $\operatorname{sg}[\cdot]$ is the stop-gradient operator and $\beta$ is the commitment weight. The forward pass uses the selected motif embedding, while the straight-through estimator propagates reconstruction gradients to the structure encoder.

We additionally discourage codebook collapse through a batch-level usage regularizer. For a mini-batch of size $B$, let$q_{im}=\operatorname{softmax}_m(-\|\mathbf h_S^{(i)}-\mathbf e_m\|_2^2/\tau_{\mathrm{div}})$ denote the soft assignment probability of structure $i$ to motif $m$, where $\tau_{\mathrm{div}}$ is the soft-assignment temperature. The batch average $\bar q_m=B^{-1}\sum_{i=1}^{B}q_{im}$ therefore represents the expected usage of motif $m$. The complete motif-learning objective is
\begin{equation}
\mathcal L_{\mathrm{motif}}
=\mathcal L_{\mathrm{VQ}}
+\lambda_{\mathrm{div}}
D_{\mathrm{KL}}\!\left(\bar{\mathbf q}\,\|\,\mathbf u_M\right),
\label{eq:motif_learning_objective}
\end{equation}
where $\mathbf u_M$ is the uniform distribution over the $M$ motifs. The regularizer encourages broader codebook utilization, while the forward path continues to use the hard nearest-neighbor assignment in Eq.~\eqref{eq:motif_quantization}. After this stage, the nearest-motif label $z^{\star}$ assigned to each active training structure is fixed and used to supervise the profile-conditioned generator.

\subsubsection{Profile-Conditioned Motif Prior}
\label{sec:motif_prior}

For an active day, the Profile-Conditioned Motif Prior predicts which high-level activity organizations are plausible for profile $\mathbf x$. A prediction head $g_z$ maps the profile representation $\mathbf h_x$ to $M$ motif logits:
\begin{equation}
P_{\theta_z}(z=m\mid\mathbf x,A=1)
=\left[\operatorname{softmax}\!\left(g_z(\mathbf h_x)\right)\right]_m,
\label{eq:motif_prior}
\end{equation}
where $m\in\{1,\ldots,M\}$. The prior is supervised by the nearest-motif label assigned by the frozen motif learner,
\begin{equation}
\mathcal L_{\mathrm{prior}}
=-\log P_{\theta_z}(z=z^{\star}\mid\mathbf x,A=1).
\label{eq:motif_prior_loss}
\end{equation}
This categorical prior allows the same profile condition to support multiple plausible activity organizations.

\subsubsection{Motif-Conditioned Exact-Structure Decoder}
\label{sec:structure_decoder}

A motif captures reusable structural information but does not determine a unique activity sequence. We therefore initialize an autoregressive GRU decoder from the profile representation $\mathbf h_x$ and motif embedding $\mathbf e_z$. During training, the decoder uses the observed motif label $z^{\star}$ and preceding activity \mbox{tokens}:
\begin{equation}
\mathcal L_{\mathrm{struct}}
=-\sum_{k=1}^{K+1}
\log P_{\theta_S}\!\left(
c_k\mid c_{0:k-1},\mathbf h_x,\mathbf e_{z^{\star}},A=1
\right),
\label{eq:structure_loss}
\end{equation}
where $c_{0:k-1}$ denotes the observed preceding tokens. The motif embedding provides shared structural information across related schedules, while the profile representation retains individual- and day-specific conditioning. At inference, adjacent identical decoded activity tokens are merged before duration allocation, yielding the canonical episode-level structure defined in Eq.~\eqref{eq:schedule_validity}.

\subsection{Autoregressive Budget-Constrained Duration Allocation}
\label{sec:duration_allocation}

Given an exact activity structure, episode durations remain stochastic but must be positive and jointly exhaust the fixed daily budget. We therefore parameterize duration allocation through a strictly increasing sequence of episode endpoints. For a structure $S=(c_1,\ldots,c_K)$, let $\tau_k$ denote the endpoint of episode $k$. We set $\tau_0=0$ and fix $\tau_K=T$, such that
\begin{equation}
0=\tau_0<\tau_1<\cdots<\tau_K=T,\qquad d_k=\tau_k-\tau_{k-1}.
\label{eq:endpoint_duration_relation}
\end{equation}
The allocator combines activity-type and episode-position embeddings and encodes the complete structure with a bidirectional GRU, producing contextual episode representations. A pooled structure representation, together with the profile representation $\mathbf h_x$ and motif embedding $\mathbf e_z$, initializes the recurrent allocation state. 

At allocation step $k$, let $\mathbf u_k$ denote the step context that combines three types of information: global profile and motif conditioning, the contextual representation of the complete activity structure, and the current allocation state, including the previous endpoint and duration, remaining time budget, and remaining episode count. The recurrent update and endpoint scoring are
\begin{equation}
\mathbf s_k=\operatorname{GRUCell}(\mathbf u_k,\mathbf s_{k-1}),\qquad
\mathbf o_k=g_{\tau}(\mathbf s_k),
\label{eq:duration_state_update}
\end{equation}
where $\mathbf s_k$ is the recurrent allocation state, $\mathbf o_k$ contains the scores of all candidate endpoint positions, and $g_{\tau}$ is the endpoint-scoring head. This context allows the allocator to account for both previous allocations and upcoming \mbox{activities}.

For each internal endpoint $\tau_k$, $k=1,\ldots,K-1$, the candidate scores $\mathbf o_k$ are masked outside the feasible range
\begin{equation}
 \tau_{k-1}+1
 \leq \tau_k
 \leq T-(K-k).
\label{eq:feasible_endpoint_range}
\end{equation}
The lower bound assigns at least one time slot to the current episode, while the upper bound reserves at least one slot for every remaining episode. A softmax over the feasible scores defines the endpoint distribution $p_k(t)$ for candidate position $t$.

During training, the observed previous endpoints and durations are used as recurrent feedback. Let $\bar\tau_k=\sum_{t=0}^{T}t\,p_k(t)$ denote the conditional mean endpoint at step $k$ under the observed preceding allocations. Setting $\bar\tau_0=0$ and $\bar\tau_K=T$, we define $\bar d_k=\bar\tau_k-\bar\tau_{k-1}$ as a boundary-derived duration surrogate for auxiliary supervision. Let $\boldsymbol\tau$ and $\bar{\boldsymbol\tau}$ collect the observed and conditional-mean internal endpoints, and let $\bar{\mathbf d}$ collect the corresponding duration surrogates. The duration objective is
\begin{equation}
\begin{aligned}
 \mathcal L_{\mathrm{dur}}
 ={}&\mathcal L_{\mathrm{end}}
 +\lambda_{\tau}\operatorname{Huber}\!\left(
 \bar{\boldsymbol\tau}/T,\boldsymbol\tau/T
 \right)\\
 &+\lambda_d\operatorname{Huber}\!\left(
 \bar{\mathbf d}/T,\mathbf d/T
 \right),
\end{aligned}
\label{eq:duration_loss}
\end{equation}
where $\mathcal L_{\mathrm{end}}=-\sum_{k=1}^{K-1}\log p_k(\tau_k)$ is the cross-entropy loss on the observed internal endpoints. The endpoint loss trains the autoregressive endpoint distributions, while the two Huber terms provide auxiliary supervision for cumulative timing and boundary-derived duration allocation. For every nonempty structure passed to the allocator, enforcing Eq.~\eqref{eq:feasible_endpoint_range} and fixing $\tau_K=T$ ensure that the sampled durations are positive integers and sum exactly to the full-day horizon, without post-hoc duration repair.

\subsection{Training Objective and Inference}
\label{sec:training_inference}

Act-HiMotif is trained in two stages. First, the motif learner is optimized on duration-free active-day structures, producing a VQ codebook and a motif label $z^{\star}$ for each active training structure. The VQ motif learner, including its codebook, and the resulting motif assignments are then frozen. In the second stage, a separate generator-side motif embedding table is initialized from the frozen codebook and optimized jointly with the remaining generation modules. For notational simplicity, $\mathbf e_z$ in the generator equations denotes this adaptive generator-side embedding, while the motif assignments $z^\star$ remain defined by the frozen VQ codebook. Thus, the discrete motif partition learned in the first stage remains fixed, while its generator-side embeddings adapt to the profile-conditioned generation objective.

The Active-Day Gate is trained with the binary cross-entropy
$\mathcal L_A$ over all person-days. The second-stage objective is
\begin{equation}
 \mathcal L
 =\lambda_A\mathcal L_A
 +\lambda_M\mathcal L_{\mathrm{prior}}
 +\lambda_S\mathcal L_{\mathrm{struct}}
 +\lambda_D\mathcal L_{\mathrm{dur}},
\label{eq:overall_training_objective}
\end{equation}
where $\mathcal L_{\mathrm{prior}}$, $\mathcal L_{\mathrm{struct}}$,
and $\mathcal L_{\mathrm{dur}}$ are evaluated only on active-day schedules. The coefficients control the relative contributions of the four training objectives. During second-stage training, the structure decoder uses the observed activity prefix, while the duration allocator uses the observed preceding endpoints and durations. At inference, the model first samples the active-day indicator. A home-only outcome directly yields $Y=((\mathrm H,T))$; otherwise, it successively samples a motif, decodes an exact activity structure, and allocates feasible episode endpoints to produce the complete schedule.

\section{Experiments}
\label{sec:experiments}

\subsection{Experimental Setup}
\label{sec:exp_setup}
\providecommand{\best}[1]{\textbf{#1}}
\providecommand{\second}[1]{\underline{#1}}
\providecommand{\improve}[1]{\textcolor{gray!70}{\textit{#1}}}

\begin{table*}
    \centering
    \caption{Five-seed mean performance under in-distribution and spatial-generalization evaluation. Best and second-best results are \textbf{bolded} and \underline{underlined}.}
    \label{tab:main_results}
    \small
    \setlength{\tabcolsep}{3.7pt}
    \renewcommand{\arraystretch}{1.08}
    \begin{tabular*}{\textwidth}{@{\extracolsep{\fill}}l*{10}{c}@{}}
        \toprule
        \multirow{2}{*}{Method}
        & \multicolumn{5}{c}{In-distribution}
        & \multicolumn{5}{c}{Spatial generalization} \\
        \cmidrule(lr){2-6}\cmidrule(lr){7-11}
        & Epi.$\downarrow$ & Dur.$\downarrow$ & Start$\downarrow$ & Slot$\downarrow$ & CPC$\uparrow$
        & Epi.$\downarrow$ & Dur.$\downarrow$ & Start$\downarrow$ & Slot$\downarrow$ & CPC$\uparrow$ \\
        \midrule
        ConditionalRNN
        & 0.3092 & 0.3646 & 0.4642 & 0.0311 & 0.4919
        & 0.3468 & 0.3571 & 0.4496 & 0.0366 & 0.4479 \\
        GenerativeRNN
        & 0.0100 & 0.0257 & 0.0334 & 0.0047 & 0.8900
        & 0.0233 & 0.0385 & 0.0512 & 0.0063 & 0.8497 \\
        ActVAE
        & 0.0314 & 0.0419 & 0.0325 & 0.0088 & 0.8591
        & 0.0459 & 0.0571 & 0.0429 & 0.0071 & 0.8770 \\
        DeepActivity-ZS
        & 0.2310 & 0.2817 & 0.3593 & 0.0269 & 0.5136
        & 0.2200 & 0.2623 & 0.3722 & 0.0282 & 0.5092 \\
        \midrule
        ControlTraj
        & 0.1492 & 0.1164 & 0.1016 & 0.0075 & 0.7184
        & 0.1810 & 0.1220 & 0.1144 & 0.0078 & 0.7278 \\
        TrajGPT
        & \second{0.0073} & \second{0.0181} & 0.0502 & \second{0.0017} & \second{0.9266}
        & \second{0.0042} & \second{0.0182} & 0.0276 & \second{0.0010} & \second{0.9338} \\
        CoDiffMob
        & 0.1373 & 0.1043 & 0.1086 & 0.0070 & 0.7116
        & 0.1468 & 0.0822 & 0.1224 & 0.0077 & 0.7285 \\
        MobilityGPT
        & 0.1490 & 0.0417 & \second{0.0144} & 0.0122 & 0.8836
        & 0.1711 & 0.0533 & \second{0.0258} & 0.0146 & 0.8891 \\
        \midrule
        \rowcolor{gray!10}
        \textbf{Act-HiMotif}
        & \best{0.0014} & \best{0.0013} & \best{0.0045} & \best{0.0001} & \best{0.9749}
        & \best{0.0021} & \best{0.0042} & \best{0.0120} & \best{0.0002} & \best{0.9653} \\
        \rowcolor{gray!4}
        \improve{Improvement}
        & \improve{80.2\%} & \improve{93.0\%} & \improve{68.9\%} & \improve{93.8\%} & \improve{5.2\%}
        & \improve{50.2\%} & \improve{76.9\%} & \improve{53.5\%} & \improve{84.8\%} & \improve{3.4\%} \\
        \bottomrule
    \end{tabular*}
    
\end{table*}

\paragraph{Dataset.}
We use the public 2017 U.S. National Household Travel Survey (NHTS), which records assigned-day travel together with demographic and household attributes~\citep{2018NHTS}. We construct 264,234 person-day schedules with nine activity categories and 48 half-hour slots, merging adjacent identical activities into episodes. The profile condition includes age band, sex, employment status, student status, education level, household-income group, and day of week. We use a person-disjoint 70/15/15 split for \emph{in-distribution} evaluation. For \emph{spatial generalization}, we use a CBSA-disjoint 70/15/15 split, where CBSAs are county-based U.S. statistical areas centered on urban cores; no disclosed CBSA overlaps across the subsets.

\paragraph{Baselines.}
We compare four activity-schedule generators---ConditionalRNN~\citep{Koushik2023ActivitySchedule}, GenerativeRNN~\citep{Shone2025DGML}, ActVAE~\citep{Shone2025ActVAE}, and DeepActivity-ZS~\citep{Liao2026DeepActivity}---and four mobility generators: ControlTraj~\citep{KDD24ControlTraj}, TrajGPT~\citep{SIGSPATIAL24TrajGPT}, CoDiffMob~\citep{WWW25CoDiffMob}, and MobilityGPT~\citep{TITS26MobilityGPT}. DeepActivity-ZS is evaluated zero-shot without target-region fine-tuning. All methods use the same splits, generation sizes, and evaluation procedures, and all conditional models receive identical profile fields.

\paragraph{Evaluation Metrics.}
We evaluate schedule complexity, temporal allocation, and sequential organization. \textbf{Episode JSD} compares episode-count distributions. \textbf{Duration JSD} and \textbf{Start JSD} compare activity-specific duration and start-time distributions and are macro-averaged over activity categories, while \textbf{Slot JSD} compares activity composition over the 48 daily slots. These metrics use Jensen--Shannon divergence~\citep{Lin1991Divergence}. \textbf{Transition CPC} measures the overlap between observed and generated activity-transition volumes~\citep{Simini2021DeepGravity,WWW25CoDiffMob}. Lower JSD and higher CPC indicate better fidelity.

\paragraph{Implementation Details.}
Act-HiMotif uses a motif codebook of size $M=256$. The motif learner and the profile-conditioned generator are trained with AdamW using a learning rate of $10^{-3}$ and a batch size of 1024 for at most 40 epochs. The second stage uses early stopping with a patience of 8 based on validation joint loss. We report results over five seeds, 2026--2030, and generate one schedule for each test profile. All experiments were conducted on a single NVIDIA GeForce RTX 3090 GPU.

Further preprocessing, baseline-adaptation, implementation, and sensitivity details are provided in Appendix~A.

\subsection{Overall Performance}
\label{sec:overall_performance}


Table~\ref{tab:main_results} compares both evaluation protocols. Among the baselines, MobilityGPT is most competitive on activity start times, whereas TrajGPT better captures schedule complexity, duration distributions, daily activity composition, and transitions. In contrast, Act-HiMotif achieves the best performance across all structural and temporal metrics. Its gains in Duration and Slot JSD indicate more faithful allocation of the daily time budget, while the improvements in Episode JSD and Transition CPC show that this temporal fidelity does not sacrifice activity-sequence organization.

These advantages persist under CBSA-disjoint evaluation, where the test regions are absent from training. The consistent improvements across both protocols indicate that Act-HiMotif captures transferable schedule regularities rather than relying on region-specific exact structures, thereby improving the fidelity of the complete schedule distribution. Full mean$\pm$standard-deviation results are reported in Appendix~B.

\subsection{Ablation Study}
\label{sec:ablation}

\begin{table}
\centering
\caption{Five-seed mean ablation results for the hierarchical components of Act-HiMotif.}
\label{tab:ablation_results}
\footnotesize
\setlength{\tabcolsep}{2.8pt}
\renewcommand{\arraystretch}{1.06}
\newcommand{\ablname}[1]{{\footnotesize #1}}
\begin{tabular}{@{}lccccc@{}}
\toprule
\multicolumn{6}{c}{\textbf{In-distribution}} \\
\midrule
Variant & Epi.$\downarrow$ & Dur.$\downarrow$ & Start$\downarrow$ & Slot$\downarrow$ & CPC$\uparrow$ \\
\midrule
\scriptsize{w/o Active-Day Gate ($A$)}         & 0.0029 & 0.0016 & 0.0047 & 0.0002 & 0.9706 \\
\scriptsize{w/o Act.-Stru. Motifs ($z$)} & 0.0223 & 0.0051 & 0.0089 & 0.0021 & 0.8910 \\
{\scriptsize
w/o Hie. Gen.
($\mathbf{x}\!\rightarrow\!Y$)}  & 0.0029 & 0.0017 & 0.0069 & 0.0007 & 0.9067 \\
\rowcolor{gray!10}
\scriptsize{\textbf{Act-HiMotif} }        & \textbf{0.0014} & \textbf{0.0013} & \textbf{0.0045} & \textbf{0.0001} & \textbf{0.9749} \\
\bottomrule
\end{tabular}

\vspace{0.45em}

\begin{tabular}{@{}lccccc@{}}
\toprule
\multicolumn{6}{c}{\textbf{Spatial generalization}} \\
\midrule
Variant & Epi.$\downarrow$ & Dur.$\downarrow$ & Start$\downarrow$ & Slot$\downarrow$ & CPC$\uparrow$ \\
\midrule
\scriptsize{w/o Active-Day Gate ($A$)}          & 0.0025 & 0.0045 & 0.0125 & 0.0003 & 0.9568 \\
\scriptsize{w/o Act.-Stru. Motifs ($z$)}   & 0.0211 & 0.0079 & 0.0175 & 0.0020 & 0.8929 \\
{\scriptsize
w/o Hie. Gen.
($\mathbf{x}\!\rightarrow\!Y$)}  & 0.0052 & \textbf{0.0022} & 0.0160 & 0.0008 & 0.9170 \\
\rowcolor{gray!10}
\scriptsize{\textbf{Act-HiMotif} }         & \textbf{0.0021} & 0.0042 & \textbf{0.0120} & \textbf{0.0002} & \textbf{0.9653} \\
\bottomrule
\end{tabular}
\end{table}

Table~\ref{tab:ablation_results} isolates Act-HiMotif's three main design choices. Removing activity-structure motifs causes the largest degradation in Episode JSD and Transition CPC under both protocols and consistently worsens temporal metrics. This indicates that, beyond compactly representing activity structures, motifs share statistical support across related activity organizations, stabilizing exact-structure generation and the structural basis for duration allocation.

Removing hierarchical generation produces a different failure mode. Although direct profile-to-schedule generation attains lower Duration JSD under spatial generalization, it substantially worsens start-time placement, daily activity composition, and transitions. Thus, matching an isolated duration marginal does not ensure schedule coherence, supporting separate structure generation and budget-constrained duration allocation. Removing the Active-Day Gate causes a smaller but consistent decline, supporting its role in separating home-only from heterogeneous active days. Full five-seed results are provided in Appendix~C.

\subsection{Further Analysis}
\label{sec:further_analysis}

\begin{figure}[!t]
    \centering
    \includegraphics[width=\linewidth]{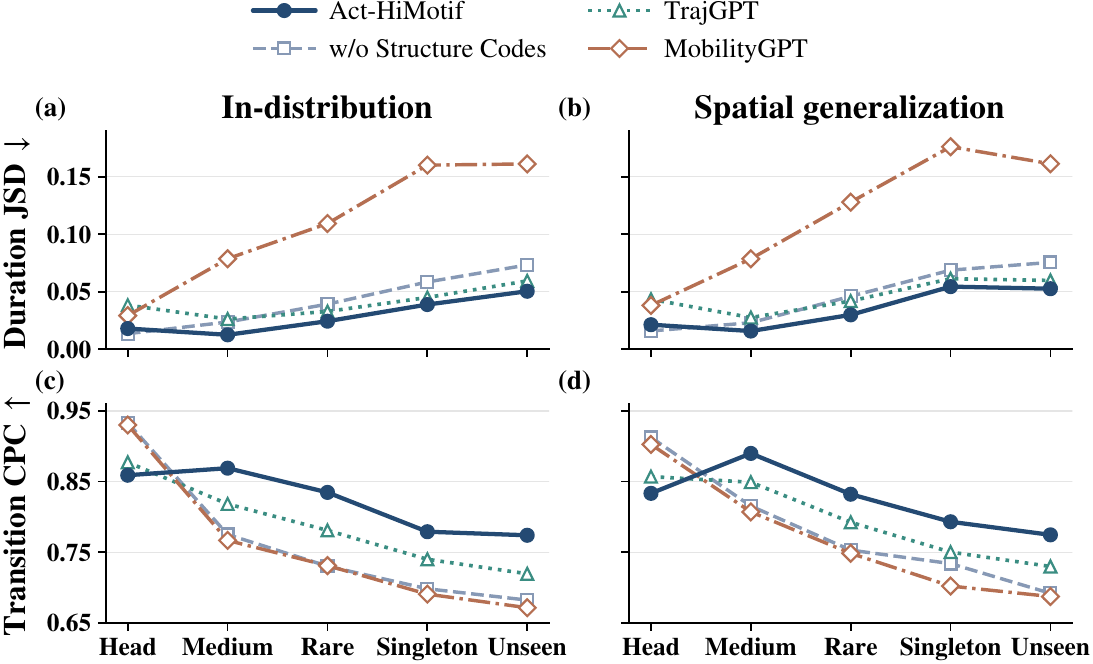}
    \caption{Duration and transition fidelity by ground-truth structure support. Groups are defined by training frequency: Head ($f_{\mathrm{train}}\geq100$), Medium ($10$--$99$), Rare ($2$--$9$), Singleton ($1$), and Unseen ($0$).}
    \label{fig:long-tailed}
\end{figure}

\paragraph{Long-Tail Structure-Support Analysis.}
Figure~\ref{fig:long-tailed} reveals a clear support-dependent pattern. On Head structures, the motif-free variant is competitive and attains higher Transition CPC, whereas Act-HiMotif maintains stronger duration fidelity. As exact-structure support decreases, Act-HiMotif consistently achieves the best Duration JSD and Transition CPC across the Medium, Rare, Singleton, and Unseen groups under both evaluation protocols. These results indicate that motif conditioning is particularly beneficial for test person-days whose ground-truth structures have limited or no exact support in the training data, consistent with its role in sharing statistical support across related activity organizations.

\begin{figure}[!t]
    \centering
    \includegraphics[width=\linewidth]{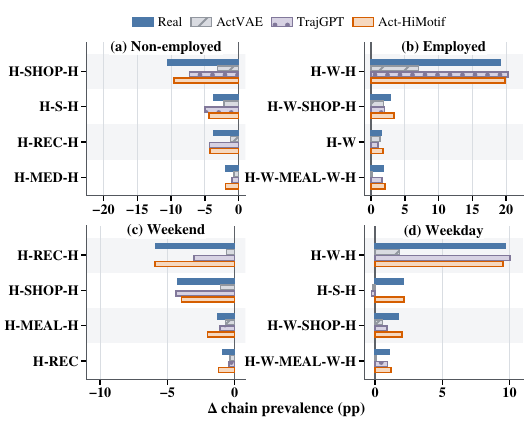}
    \caption{Profile--activity-chain alignment under CBSA-disjoint spatial generalization. The top four active-day chains by training-set prevalence difference for the corresponding group.}
    \label{fig:profile_chain_alignment}
\end{figure}

\paragraph{Profile--Activity-Chain Alignment.}
Figure~\ref{fig:profile_chain_alignment} examines whether profile conditioning preserves profile-specific preferences over exact activity structures. For employment and day type, the figure shows the four active-day chains with the largest training-set prevalence differences toward each profile group. Under CBSA-disjoint spatial generalization, Act-HiMotif generally follows the real contrasts across all four groups, preserving non-employed versus employed and weekend versus weekday preferences. Competing generators more often attenuate these differences toward zero or distort their relative strengths. This indicates that Act-HiMotif better retains profile-specific preferences over exact activity structures. Full chain-alignment and subgroup-level profile analyses are provided in Appendix~D.

\begin{figure}[!t]
    \centering
    \includegraphics[width=\linewidth]{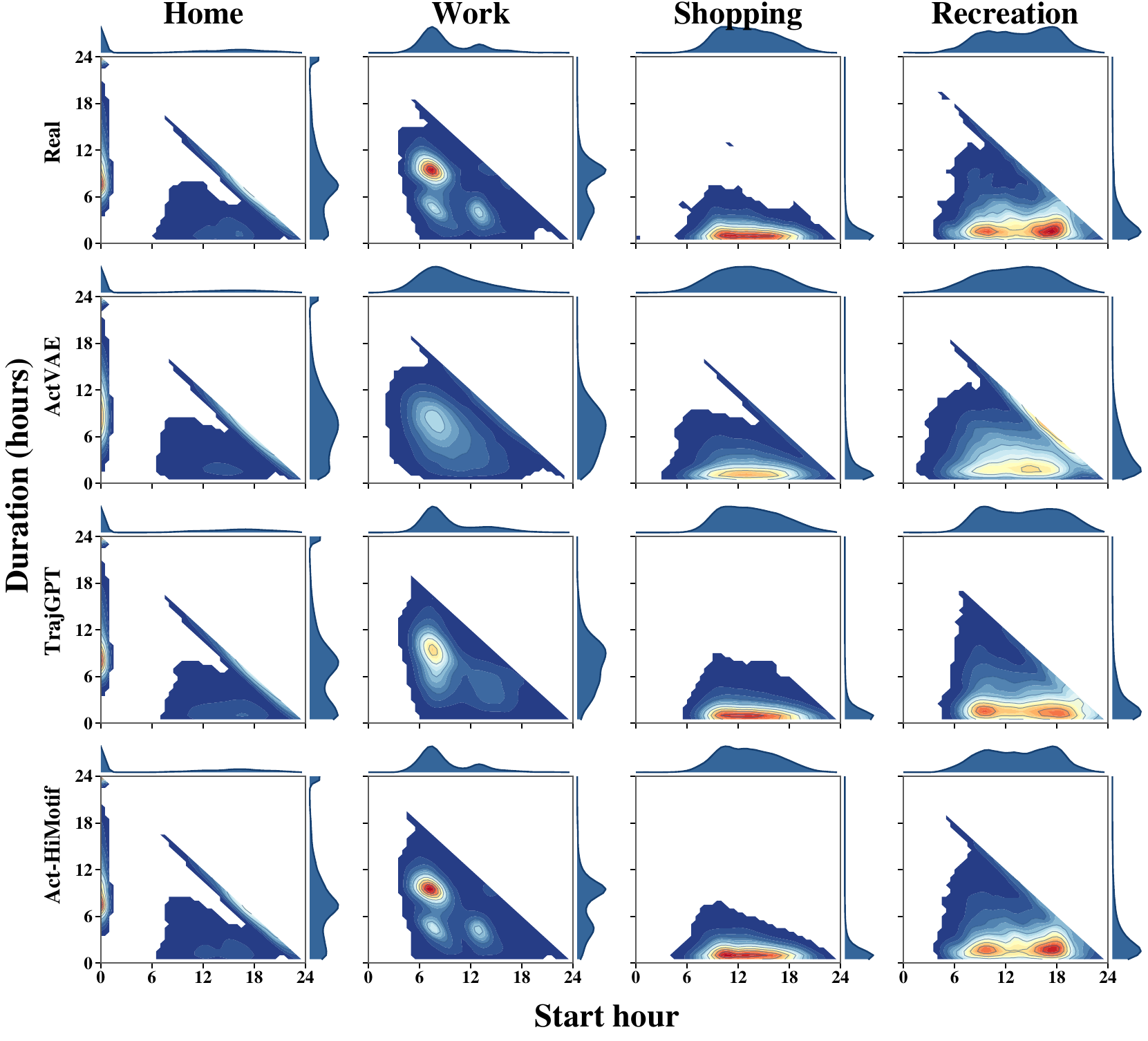}
    \caption{Observed and generated joint start-time--duration distributions for four representative activities under CBSA-disjoint spatial generalization.}
    \label{fig:start_duration_joint}
\end{figure}

\paragraph{Structural and Joint Temporal Fidelity.}
Marginal Start and Duration JSD do not reveal whether activity durations occur at appropriate times of day. Figure~\ref{fig:start_duration_joint} therefore compares their joint distributions under CBSA-disjoint spatial generalization. Across Home, Work, Shopping, and Recreation, Act-HiMotif more closely matches the observed joint temporal patterns than ActVAE and TrajGPT, preserving the broad Home distribution, the concentrated morning-start, long-duration Work mode, and the distinct patterns of Shopping and Recreation. Together with its strong Transition CPC in Table~\ref{tab:main_results}, these results indicate that Act-HiMotif captures start--duration dependence while maintaining sequential organization. Additional activity-wise results, transition matrices, and in-distribution analyses are reported in Appendix~E.

\section{Conclusion}
\label{sec:conclusion}

Act-HiMotif addresses profile-conditioned daily activity schedule generation through a hierarchical motif-based framework that introduces learned, reusable activity-structure motifs between profile conditions and exact activity structures, followed by budget-constrained duration allocation. The results show that activity-structure motifs are especially beneficial for exact structures with limited training support, while separating structure generation from budget-constrained duration allocation improves complete-schedule coherence. More broadly, profile-conditioned generators can assign diverse schedules to synthetic population profiles that can be constructed from demographic statistics and microdata such as ACS and PUMS \citep{USCensus2024ACSMethodology, USCensus2021PUMSHandbook}, providing a practical route to population-scale activity-schedule datasets for computational social science, human behavior simulation, activity-based travel modeling, and other profile-aware downstream applications.

\noindent\textbf{Generative AI Use Statement.}
Large language models were used solely for language polishing. The authors reviewed the final manuscript and take full responsibility for its content.

\bibliography{reference}

@misc{Shone2025ActVAE,
      title={Black box behavioural modelling: Predicting human activity schedules with a deep conditional generative approach}, 
      author={Fred Shone and Tim Hillel},
      year={2026},
      eprint={2512.04223},
      archivePrefix={arXiv},
      primaryClass={cs.LG},
      url={https://arxiv.org/abs/2512.04223}, 
}

@article{Shone2025DGML,
  author = {Shone, Fred and Hillel, Tim},
  title = {Synthesising Activity Participations and Scheduling with Deep Generative Machine Learning},
  journal = {Transportation Research Part C: Emerging Technologies},
  volume = {179},
  pages = {105273},
  year = {2025},
  doi = {10.1016/j.trc.2025.105273},
  url = {https://www.sciencedirect.com/science/article/pii/S0968090X25002773}
}

@article{Liao2026DeepActivity,
  author = {Liao, Xishun and Jiang, Qinhua and He, Brian Yueshuai and Liu, Yifan and Kuai, Chenchen and Ma, Jiaqi},
  title = {Deep Activity Model: A Generative Deep Learning Approach for Human Mobility Pattern Synthesis},
  journal = {IEEE Transactions on Intelligent Transportation Systems},
  volume = {27},
  number = {6},
  pages = {6519--6536},
  year = {2026},
  doi = {10.1109/TITS.2026.3678542}
}

@inproceedings{vanDenOord2017VQVAE,
 author = {van den Oord, Aaron and Vinyals, Oriol and kavukcuoglu, koray},
 booktitle = {Advances in Neural Information Processing Systems},
 editor = {I. Guyon and U. Von Luxburg and S. Bengio and H. Wallach and R. Fergus and S. Vishwanathan and R. Garnett},
 pages = {},
 publisher = {Curran Associates, Inc.},
 title = {Neural Discrete Representation Learning},
 url = {https://proceedings.neurips.cc/paper_files/paper/2017/file/7a98af17e63a0ac09ce2e96d03992fbc-Paper.pdf},
 volume = {30},
 year = {2017}
}

@misc{yang2026activityeditor,
      title={ActivityEditor: Learning to Synthesize Physically Valid Human Mobility}, 
      author={Chenjie Yang and Yutian Jiang and Anqi Liang and Wei Qi and Chenyu Wu and Junbo Zhang},
      year={2026},
      eprint={2604.05529},
      archivePrefix={arXiv},
      primaryClass={cs.AI},
      url={https://arxiv.org/abs/2604.05529}, 
}

@misc{Li2024MobAgent,
      title={Be More Real: Travel Diary Generation Using LLM Agents and Individual Profiles}, 
      author={Xuchuan Li and Fei Huang and Jianrong Lv and Zhixiong Xiao and Guolong Li and Yang Yue},
      year={2024},
      eprint={2407.18932},
      archivePrefix={arXiv},
      primaryClass={cs.CY},
      url={https://arxiv.org/abs/2407.18932}, 
}

@article{Bowman2001ActivityBased,
  author = {Bowman, J. L. and Ben-Akiva, M. E.},
  title = {Activity-Based Disaggregate Travel Demand Model System with Activity Schedules},
  journal = {Transportation Research Part A: Policy and Practice},
  volume = {35},
  number = {1},
  pages = {1--28},
  year = {2001},
  doi = {10.1016/S0965-8564(99)00043-9}
}

@article{Arentze2004ALBATROSS,
  author = {Arentze, Theo A. and Timmermans, Harry J. P.},
  title = {A Learning-Based Transportation Oriented Simulation System},
  journal = {Transportation Research Part B: Methodological},
  volume = {38},
  number = {7},
  pages = {613--633},
  year = {2004},
  doi = {10.1016/j.trb.2002.10.001}
}

@inproceedings{Wang2024LLMob,
  author = {Wang, Jiawei and Jiang, Renhe and Yang, Chuang and Wu, Zengqing and Onizuka, Makoto and Shibasaki, Ryosuke and Koshizuka, Noboru and Xiao, Chuan},
  title = {Large Language Models as Urban Residents: An {LLM} Agent Framework for Personal Mobility Generation},
  booktitle = {Advances in Neural Information Processing Systems},
  volume = {37},
  pages = {124547--124574},
  year = {2024},
  publisher = {Neural Information Processing Systems Foundation},
  doi = {10.52202/079017-3957}
}

@article{Bhat2008MDCEV,
  author = {Bhat, Chandra R.},
  title = {The Multiple Discrete-Continuous Extreme Value ({MDCEV}) Model: Role of Utility Function Parameters, Identification Considerations, and Model Extensions},
  journal = {Transportation Research Part B: Methodological},
  volume = {42},
  number = {3},
  pages = {274--303},
  year = {2008},
  doi = {10.1016/j.trb.2007.06.002}
}

@article{Koushik2023ActivitySchedule,
  author = {Koushik, Anil and Manoj, M. and Nezamuddin, N. and Prathosh, A. P.},
  title = {Activity Schedule Modeling Using Machine Learning},
  journal = {Transportation Research Record: Journal of the Transportation Research Board},
  volume = {2677},
  number = {8},
  pages = {1--23},
  year = {2023},
  doi = {10.1177/03611981231155426}
}

@inproceedings{Liu2026TravelReasoner,
  author = {Liu, Peijie and Xu, Fengli and Li, Yong},
  title = {{TravelReasoner}: Leveraging Large Reasoning Models to Address Mobility Data Gap},
  booktitle = {Proceedings of the ACM Web Conference 2026},
  series = {WWW '26},
  pages = {9341--9351},
  year = {2026},
  publisher = {Association for Computing Machinery},
  address = {New York, NY, USA},
  isbn = {9798400723070},
  doi = {10.1145/3774904.3793054}
}

@article{Luca2021MobilitySurvey,
  author = {Luca, Massimiliano and Barlacchi, Gianni and Lepri, Bruno and Pappalardo, Luca},
  title = {A Survey on Deep Learning for Human Mobility},
  journal = {ACM Computing Surveys},
  volume = {55},
  number = {1},
  pages = {1--44},
  year = {2021},
  doi = {10.1145/3485125}
}

@article{Krishna2018ActivityDuration,
  author = {Krishna, Kundan and Jain, Deepali and Mehta, Sanket V. and Choudhary, Sunav},
  title = {An {LSTM}-Based System for Prediction of Human Activities with Durations},
  journal = {Proceedings of the ACM on Interactive, Mobile, Wearable and Ubiquitous Technologies},
  volume = {1},
  number = {4},
  pages = {1--31},
  year = {2018},
  doi = {10.1145/3161201}
}

@article{Yuan2024NeedDynamics,
  author = {Yuan, Yuan and Ding, Jingtao and Wang, Huandong and Jin, Depeng},
  title = {Generating Daily Activities with Need Dynamics},
  journal = {ACM Transactions on Intelligent Systems and Technology},
  volume = {15},
  number = {2},
  pages = {1--28},
  year = {2024},
  doi = {10.1145/3637493}
}

@misc{BaduMarfo2020DPDiary,
  author = {Badu-Marfo, Godwin and Farooq, Bilal and Patterson, Zachary},
  title = {A Differentially Private Multi-Output Deep Generative Networks Approach for Activity Diary Synthesis},
  year = {2020},
  eprint = {2012.14574},
  archivePrefix = {arXiv},
  primaryClass = {cs.LG}
}

@inproceedings{Yuan2022ActivityTrajectory,
  author = {Yuan, Yuan and Ding, Jingtao and Wang, Huandong and Jin, Depeng and Li, Yong},
  title = {Activity Trajectory Generation via Modeling Spatiotemporal Dynamics},
  booktitle = {Proceedings of the 28th ACM SIGKDD Conference on Knowledge Discovery and Data Mining},
  series = {KDD '22},
  pages = {4752--4762},
  year = {2022},
  publisher = {Association for Computing Machinery},
  address = {New York, NY, USA},
  doi = {10.1145/3534678.3542671}
}

@article{Kapp2024SyntheticMobilitySurvey,
author = {Kapp, Alexandra and Hansmeyer, Julia and Mihaljevi\'{c}, Helena},
title = {Generative Models for Synthetic Urban Mobility Data: A Systematic Literature Review},
year = {2023},
issue_date = {April 2024},
publisher = {Association for Computing Machinery},
address = {New York, NY, USA},
volume = {56},
number = {4},
issn = {0360-0300},
url = {https://doi.org/10.1145/3610224},
doi = {10.1145/3610224},
journal = {ACM Comput. Surv.},
month = nov,
articleno = {93},
numpages = {37}
}

@article{Liu2024Act2Loc,
author = {Kang Liu and Xin Jin and Shifen Cheng and Song Gao and Ling Yin and Feng Lu},
title = {Act2Loc: a synthetic trajectory generation method by combining machine learning and mechanistic models},
journal = {International Journal of Geographical Information Science},
volume = {38},
number = {3},
pages = {407--431},
year = {2024},
publisher = {Taylor \& Francis},
doi = {10.1080/13658816.2023.2292570},
URL = {https://doi.org/10.1080/13658816.2023.2292570},
eprint = {https://doi.org/10.1080/13658816.2023.2292570}
}

@inproceedings{Park2023GenerativeAgents,
author = {Park, Joon Sung and O'Brien, Joseph and Cai, Carrie Jun and Morris, Meredith Ringel and Liang, Percy and Bernstein, Michael S.},
title = {Generative Agents: Interactive Simulacra of Human Behavior},
year = {2023},
isbn = {9798400701320},
publisher = {Association for Computing Machinery},
address = {New York, NY, USA},
url = {https://doi.org/10.1145/3586183.3606763},
doi = {10.1145/3586183.3606763},
booktitle = {Proceedings of the 36th Annual ACM Symposium on User Interface Software and Technology},
articleno = {2},
numpages = {22},
location = {San Francisco, CA, USA},
series = {UIST '23}
}

@article{Pougala2023OASIS,
title = {OASIS: Optimisation-based Activity Scheduling with Integrated Simultaneous choice dimensions},
journal = {Transportation Research Part C: Emerging Technologies},
volume = {155},
pages = {104291},
year = {2023},
issn = {0968-090X},
doi = {https://doi.org/10.1016/j.trc.2023.104291},
url = {https://www.sciencedirect.com/science/article/pii/S0968090X23002802},
author = {Janody Pougala and Tim Hillel and Michel Bierlaire}
}

@inproceedings{WWW25CoDiffMob,
author = {Zhang, Yuheng and Yuan, Yuan and Ding, Jingtao and Yuan, Jian and Li, Yong},
title = {Noise Matters: Diffusion Model-based Urban Mobility Generation with Collaborative Noise Priors},
year = {2025},
isbn = {9798400712746},
publisher = {Association for Computing Machinery},
address = {New York, NY, USA},
url = {https://doi.org/10.1145/3696410.3714516},
doi = {10.1145/3696410.3714516},
booktitle = {Proceedings of the ACM on Web Conference 2025},
pages = {5352–5363},
numpages = {12},
location = {Sydney NSW, Australia},
series = {WWW '25}
}

@inproceedings{KDD24ControlTraj,
author = {Zhu, Yuanshao and Yu, James Jianqiao and Zhao, Xiangyu and Liu, Qidong and Ye, Yongchao and Chen, Wei and Zhang, Zijian and Wei, Xuetao and Liang, Yuxuan},
title = {ControlTraj: Controllable Trajectory Generation with Topology-Constrained Diffusion Model},
year = {2024},
isbn = {9798400704901},
publisher = {Association for Computing Machinery},
address = {New York, NY, USA},
url = {https://doi.org/10.1145/3637528.3671866},
doi = {10.1145/3637528.3671866},
booktitle = {Proceedings of the 30th ACM SIGKDD Conference on Knowledge Discovery and Data Mining},
pages = {4676–4687},
numpages = {12},
location = {Barcelona, Spain},
series = {KDD '24}
}

@ARTICLE{TITS26MobilityGPT,
  author={Haydari, Ammar and Chen, Dongjie and Lai, Zhengfeng and Zhang, Michael and Chuah, Chen-Nee},
  journal={IEEE Transactions on Intelligent Transportation Systems}, 
  title={MobilityGPT: Enhanced Human Mobility Modeling With a GPT Model}, 
  year={2026},
  volume={27},
  number={1},
  pages={1681-1694},
  doi={10.1109/TITS.2025.3626357}}

@inproceedings{SIGSPATIAL24TrajGPT,
author = {Hsu, Shang-Ling and Tung, Emmanuel and Krumm, John and Shahabi, Cyrus and Shafique, Khurram},
title = {TrajGPT: Controlled Synthetic Trajectory Generation Using a Multitask Transformer-Based Spatiotemporal Model},
year = {2024},
isbn = {9798400711077},
publisher = {Association for Computing Machinery},
address = {New York, NY, USA},
url = {https://doi.org/10.1145/3678717.3691303},
doi = {10.1145/3678717.3691303},
booktitle = {Proceedings of the 32nd ACM International Conference on Advances in Geographic Information Systems},
pages = {362–371},
numpages = {10},
location = {Atlanta, GA, USA},
series = {SIGSPATIAL '24}
}

@techreport{2018NHTS,
  author      = {McGuckin, N. and Fucci, A.},
  title       = {Summary of Travel Trends: {2017 National Household Travel Survey}},
  institution = {Federal Highway Administration,
                 Office of Policy and Governmental Affairs},
  number      = {FHWA-PL-18-019},
  year        = {2018},
  month       = jul,
  address     = {Washington, DC},
  url         = {https://rosap.ntl.bts.gov/view/dot/68751},
  note        = {Prepared by Westat, Inc.}
}

@article{Lin1991Divergence,
  author  = {Lin, Jianhua},
  title   = {Divergence Measures Based on the Shannon Entropy},
  journal = {IEEE Transactions on Information Theory},
  volume  = {37},
  number  = {1},
  pages   = {145--151},
  year    = {1991},
  doi     = {10.1109/18.61115}
}

@article{Simini2021DeepGravity,
  author  = {Simini, Filippo and Barlacchi, Gianni and Luca, Massimiliano and Pappalardo, Luca},
  title   = {A Deep Gravity Model for Mobility Flows Generation},
  journal = {Nature Communications},
  volume  = {12},
  pages   = {6576},
  year    = {2021},
  doi     = {10.1038/s41467-021-26752-4}
}

@article{Drchal2019DataDrivenScheduler,
title = {Data-driven activity scheduler for agent-based mobility models},
journal = {Transportation Research Part C: Emerging Technologies},
volume = {98},
pages = {370-390},
year = {2019},
issn = {0968-090X},
doi = {https://doi.org/10.1016/j.trc.2018.12.002},
url = {https://www.sciencedirect.com/science/article/pii/S0968090X18306417},
author = {Jan Drchal and Michal Čertický and Michal Jakob}
}

@article{Schneider2013MobilityMotifs,
  author = {Schneider, Christian M. and Belik, Vitaly and Couronn{\'e}, Thomas and Smoreda, Zbigniew and Gonz{\'a}lez, Marta C.},
  title = {Unravelling Daily Human Mobility Motifs},
  journal = {Journal of the Royal Society Interface},
  volume = {10},
  number = {84},
  pages = {20130246},
  year = {2013},
  doi = {10.1098/rsif.2013.0246}
}

@inproceedings{Xu2024LongTailMobility,
  author = {Xu, Xiaohang and Jiang, Renhe and Yang, Chuang and Fan, Zipei and Sezaki, Kaoru},
  title = {Taming the Long Tail in Human Mobility Prediction},
  booktitle = {Advances in Neural Information Processing Systems},
  volume = {37},
  pages = {54748--54771},
  year = {2024},
  doi = {10.52202/079017-1736}
}

@techreport{USCensus2024ACSMethodology,
  author      = {{U.S. Census Bureau}},
  title       = {American Community Survey and Puerto Rico
                 Community Survey Design and Methodology},
  institution = {U.S. Census Bureau},
  address     = {Washington, DC},
  year        = {2024},
  month       = dec,
  number      = {Version 4.0},
  url         = {https://www2.census.gov/programs-surveys/acs/
                 methodology/design_and_methodology/2024/
                 acs_design_methodology_report_2024.pdf}
}

@manual{USCensus2021PUMSHandbook,
  author       = {{U.S. Census Bureau}},
  title        = {Understanding and Using the American Community
                  Survey Public Use Microdata Sample Files:
                  What Data Users Need to Know},
  organization = {U.S. Census Bureau},
  address      = {Washington, DC},
  year         = {2021},
  month        = feb,
  url          = {https://www.census.gov/content/dam/Census/
                  library/publications/2021/acs/
                  acs_pums_handbook_2021.pdf}
}

@article{LYU2026FLAIR,
title = {FLAIR: An LLM-integrated multi-layer framework for fine-grained activity type inference in human trajectory data},
journal = {Transportation Research Part C: Emerging Technologies},
volume = {190},
pages = {105787},
year = {2026},
issn = {0968-090X},
doi = {https://doi.org/10.1016/j.trc.2026.105787},
url = {https://www.sciencedirect.com/science/article/pii/S0968090X26002755},
author = {Yilin Lyu and Kang Liu and Zhongcai Cao and Yuxiao Luo and Ling Yin and Tengfei Yang and Zugang Chen}
}

@article{Fan2026UrbanMixing,
  title   = {Hidden Patterns of Urban Mixing across Five Global Cities},
  author  = {Fan, Zhuangyuan and Loo, Becky P. Y. and Duarte, Fabio and Ratti, Carlo and Moro, Esteban},
  journal = {Nature Cities},
  year    = {2026},
  doi     = {10.1038/s44284-026-00477-1}
}

\clearpage
\twocolumn[{\centering\LARGE\bfseries Appendix\par\vspace{1em}}]

\section{A. Experimental Details}
\label{app:experimental_details}

\subsection{Data Processing and Dataset Statistics}

Person-day schedules are constructed from the public person and trip records of the 2017 U.S. National Household Travel Survey (NHTS)~\citep{2018NHTS}. Trip records are ordered by departure time within each assigned survey day. Time before the first recorded departure is labeled as Home, and each subsequent interval is assigned the destination-purpose activity of the corresponding trip. Travel is not represented as a separate activity category; each interval following a recorded trip is labeled by the destination-purpose activity associated with that trip. Person-days without a valid trip departure are retained as home-only schedules. Adjacent episodes with the same activity label are merged.

Table~\ref{tab:dataset_summary} summarizes the processed dataset statistics and the exact numbers of person-days used under the two evaluation protocols.

\begin{table}[h]
    \centering
    \caption{Processed dataset statistics and evaluation splits.}
    \label{tab:dataset_summary}
    \footnotesize
    \setlength{\tabcolsep}{4pt}
    \renewcommand{\arraystretch}{1.06}
    \begin{threeparttable}
    \begin{tabularx}{\columnwidth}{@{}Xrrr@{}}
        \toprule
        \multicolumn{4}{@{}l}{\textit{Dataset statistics}} \\
        Statistic & \multicolumn{3}{r}{Value} \\
        \midrule
        Person-day schedules
            & \multicolumn{3}{r}{264,234} \\
        Activity categories
            & \multicolumn{3}{r}{9} \\
        Temporal resolution
            & \multicolumn{3}{r}{48 slots (30 min)} \\
        Profile fields
            & \multicolumn{3}{r}{7} \\
        Distinct exact activity structures
            & \multicolumn{3}{r}{20,425} \\
        Distinct complete schedules
            & \multicolumn{3}{r}{109,669} \\
        \midrule
        \multicolumn{4}{@{}l}{\textit{Evaluation splits (person-days)}} \\
        Protocol & Train & Val. & Test \\
        \midrule
        In-distribution
            & 184,964 & 39,635 & 39,635 \\
        Spatial generalization
            & 81,426 & 17,408 & 17,394 \\
        \bottomrule
    \end{tabularx}
    \begin{tablenotes}[flushleft]
        \footnotesize
        \item 148,006 person-days without a disclosed CBSA are excluded
        from the spatial-generalization protocol.
    \end{tablenotes}
    \end{threeparttable}
\end{table}

Categorical vocabularies are constructed exclusively from the corresponding training partition, with missing or unseen values mapped to an unknown category. Validation partitions are used for model selection, and test partitions are reserved for final generation and evaluation.

\subsection{Baseline Adaptation}

Each baseline is adapted to the common 48-slot activity-schedule representation while retaining its recurrent, variational, autoregressive, or diffusion-based generation mechanism. ConditionalRNN~\citep{Koushik2023ActivitySchedule}, GenerativeRNN~\citep{Shone2025DGML}, ActVAE~\citep{Shone2025ActVAE}, and DeepActivity~\citep{Liao2026DeepActivity} generate daily activity
schedules directly. TrajGPT~\citep{SIGSPATIAL24TrajGPT} and MobilityGPT~\citep{TITS26MobilityGPT} generate semantic activity tokens autoregressively, while ControlTraj~\citep{KDD24ControlTraj} and
CoDiffMob~\citep{WWW25CoDiffMob} generate categorical activity representations through diffusion-based models. Activity vocabularies and any transition or topology information required by a baseline are estimated exclusively from its training partition.

DeepActivity-FT20 uses 20\% of the target-CBSA schedules for adaptation and is evaluated on the remaining target schedules. Its results are reported separately from methods that do not access target-CBSA schedules.

\subsection{Implementation Details}

The motif learner uses 64-dimensional activity and motif representations and a bidirectional GRU with hidden size 128. The commitment weight is set to $\beta=0.25$, while the motif-usage regularizer uses $\lambda_{\mathrm{div}}=0.2$ and $\tau_{\mathrm{div}}=0.5$. The motif checkpoint is selected according to validation reconstruction loss.

The Profile Encoder and Structure Decoder use hidden dimensions of 128, and the Duration Allocator uses a hidden dimension of 192. The second-stage loss weights are $(\lambda_A,\lambda_M,\lambda_S,\lambda_D)=(0.2,1,1,1)$, while the auxiliary endpoint and duration weights are $\lambda_{\tau}=\lambda_d=0.1$. At inference, the motif, structure, and duration sampling temperatures are set to $1.0$, $0.8$, and $1.0$, respectively. These settings are fixed across evaluation protocols and random seeds.

\subsection{Hyperparameter Sensitivity}

We conduct one-at-a-time sensitivity analyses for the codebook size $M$, commitment weight $\beta$, and second-stage loss weights $\lambda_A$, $\lambda_M$, $\lambda_S$, and $\lambda_D$. When one hyperparameter is varied, all others are fixed at $M=256$, $\beta=0.25$, $\lambda_A=0.2$, and $\lambda_M=\lambda_S=\lambda_D=1$.

\begin{figure*}[!t]
    \centering
    \includegraphics[width=\textwidth]{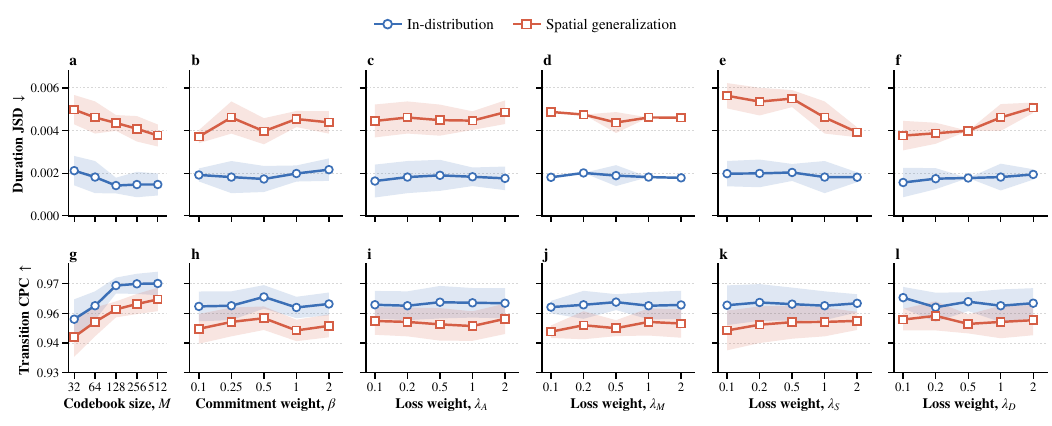}
    \caption{One-at-a-time sensitivity of Duration JSD and Transition CPC to the codebook size $M$, commitment weight $\beta$, and loss weights $\lambda_A$, $\lambda_M$, $\lambda_S$, and $\lambda_D$. Lines report five-seed means, and shaded regions indicate $\pm1$ standard deviation separately under each evaluation protocol.}
    \label{fig:parameter_sensitivity}
\end{figure*}

Figure~\ref{fig:parameter_sensitivity} reports the resulting Duration JSD and Transition CPC under both evaluation protocols. The codebook size exhibits the clearest capacity effect: increasing $M$ from 32 to
256 improves both metrics, while the additional gains at $M=512$ are limited. The remaining hyperparameters produce moderate variations around their default values, with the selected configuration lying in a relatively stable region under both protocols.

\section{B. Overall Results}
\label{app:overall_results}

Table~\ref{tab:full_results_appendix} reports mean$\pm$standard deviation over five seeds for all compared methods. DeepActivity-FT20 is presented in a separate target-adapted block.


\providecommand{\best}[1]{\textbf{#1}}
\providecommand{\second}[1]{\underline{#1}}
\providecommand{\ms}[2]{#1\,{\scriptsize$\pm$\,#2}}

\begin{table*}[ht]
    \centering
    \caption{Complete five-seed results under in-distribution and spatial-generalization evaluation. Values are mean$\pm$standard deviation. Best and second-best results among methods using no target-CBSA schedules are shown in \textbf{bold} and \underline{underline}, respectively.}
    \label{tab:full_results_appendix}

    \scriptsize
    \setlength{\tabcolsep}{2.15pt}
    \renewcommand{\arraystretch}{1.08}

    \resizebox{\textwidth}{!}{%
    \begin{tabular}{@{}lc*{10}{c}@{}}
        \toprule
        \multirow{2}{*}{Method}
        & \multirow{2}{*}{\shortstack{Target\\data}}
        & \multicolumn{5}{c}{In-distribution}
        & \multicolumn{5}{c}{Spatial generalization} \\
        \cmidrule(lr){3-7}\cmidrule(lr){8-12}
        &
        & Epi.$\downarrow$ & Dur.$\downarrow$ & Start$\downarrow$ & Slot$\downarrow$ & CPC$\uparrow$
        & Epi.$\downarrow$ & Dur.$\downarrow$ & Start$\downarrow$ & Slot$\downarrow$ & CPC$\uparrow$ \\
        \midrule
        \multicolumn{12}{@{}l}{\textit{No target-CBSA schedules}} \\
        \addlinespace[1pt]
        ConditionalRNN
        & 0\%
        & \ms{0.3092}{0.0074} & \ms{0.3646}{0.0199} & \ms{0.4642}{0.0179} & \ms{0.0311}{0.0035} & \ms{0.4919}{0.0256}
        & \ms{0.3468}{0.0076} & \ms{0.3571}{0.0314} & \ms{0.4496}{0.0321} & \ms{0.0366}{0.0046} & \ms{0.4479}{0.0270} \\
        GenerativeRNN
        & 0\%
        & \ms{0.0100}{0.0021} & \ms{0.0257}{0.0057} & \ms{0.0334}{0.0053} & \ms{0.0047}{0.0019} & \ms{0.8900}{0.0207}
        & \ms{0.0233}{0.0100} & \ms{0.0385}{0.0094} & \ms{0.0512}{0.0129} & \ms{0.0063}{0.0033} & \ms{0.8497}{0.0263} \\
        ActVAE
        & 0\%
        & \ms{0.0314}{0.0075} & \ms{0.0419}{0.0097} & \ms{0.0325}{0.0027} & \ms{0.0088}{0.0022} & \ms{0.8591}{0.0125}
        & \ms{0.0459}{0.0059} & \ms{0.0571}{0.0058} & \ms{0.0429}{0.0046} & \ms{0.0071}{0.0032} & \ms{0.8770}{0.0288} \\
        DeepActivity-ZS
        & 0\%
        & \ms{0.2310}{0.0427} & \ms{0.2817}{0.0436} & \ms{0.3593}{0.0205} & \ms{0.0269}{0.0045} & \ms{0.5136}{0.0396}
        & \ms{0.2200}{0.0440} & \ms{0.2623}{0.0392} & \ms{0.3722}{0.0104} & \ms{0.0282}{0.0045} & \ms{0.5092}{0.0692} \\
        \midrule
        ControlTraj
        & 0\%
        & \ms{0.1492}{0.0100} & \ms{0.1164}{0.0068} & \ms{0.1016}{0.0032} & \ms{0.0075}{0.0010} & \ms{0.7184}{0.0217}
        & \ms{0.1810}{0.0175} & \ms{0.1220}{0.0132} & \ms{0.1144}{0.0067} & \ms{0.0078}{0.0006} & \ms{0.7278}{0.0174} \\
        TrajGPT
        & 0\%
        & \second{\ms{0.0073}{0.0038}} & \second{\ms{0.0181}{0.0124}} & \ms{0.0502}{0.0285} & \second{\ms{0.0017}{0.0014}} & \second{\ms{0.9266}{0.0242}}
        & \second{\ms{0.0042}{0.0014}} & \second{\ms{0.0182}{0.0111}} & \ms{0.0276}{0.0057} & \second{\ms{0.0010}{0.0004}} & \second{\ms{0.9338}{0.0121}} \\
        CoDiffMob
        & 0\%
        & \ms{0.1373}{0.0128} & \ms{0.1043}{0.0064} & \ms{0.1086}{0.0051} & \ms{0.0070}{0.0006} & \ms{0.7116}{0.0211}
        & \ms{0.1468}{0.0030} & \ms{0.0822}{0.0104} & \ms{0.1224}{0.0106} & \ms{0.0077}{0.0008} & \ms{0.7285}{0.0196} \\
        MobilityGPT
        & 0\%
        & \ms{0.1490}{0.0041} & \ms{0.0417}{0.0030} & \second{\ms{0.0144}{0.0011}} & \ms{0.0122}{0.0011} & \ms{0.8836}{0.0082}
        & \ms{0.1711}{0.0058} & \ms{0.0533}{0.0032} & \second{\ms{0.0258}{0.0026}} & \ms{0.0146}{0.0013} & \ms{0.8891}{0.0074} \\
        \midrule
        \rowcolor{gray!10}
        \textbf{Act-HiMotif}
        & 0\%
        & \best{\ms{0.0014}{0.0007}} & \best{\ms{0.0013}{0.0001}} & \best{\ms{0.0045}{0.0003}} & \best{\ms{0.0001}{0.0001}} & \best{\ms{0.9749}{0.0027}}
        & \best{\ms{0.0021}{0.0009}} & \best{\ms{0.0042}{0.0003}} & \best{\ms{0.0120}{0.0010}} & \best{\ms{0.0002}{0.0001}} & \best{\ms{0.9653}{0.0038}} \\
        \midrule
        \multicolumn{12}{@{}l}{\textit{Target-adapted setting}} \\
        \addlinespace[1pt]
        DeepActivity-FT20
        & 20\%
        & \multicolumn{5}{c}{---}
        & \ms{0.1961}{0.0639} & \ms{0.2206}{0.0292} & \ms{0.3513}{0.0229} & \ms{0.0249}{0.0053} & \ms{0.5663}{0.0743} \\
        \bottomrule
    \end{tabular}%
    }
\end{table*}

\section{C. Ablation and Model Analyses}
\label{app:model_analysis}

\subsection{Full Ablation Results}

Table~\ref{tab:ablation_full_results} reports mean$\pm$standard deviation over five seeds for the three ablation variants and the full Act-HiMotif model.


\providecommand{\best}[1]{\textbf{#1}}
\providecommand{\ms}[2]{#1\,{\scriptsize$\pm$\,#2}}

\begin{table*}[ht]
\centering
\begin{threeparttable}
\caption{Full five-seed ablation results reported as mean$\pm$standard deviation.}
\label{tab:ablation_full_results}
\scriptsize
\setlength{\tabcolsep}{4.8pt}
\renewcommand{\arraystretch}{1.08}

\begin{tabular*}{\textwidth}{@{\extracolsep{\fill}}lccccc@{}}
\toprule
\multicolumn{6}{c}{\textbf{In-distribution}} \\
\cmidrule(lr){1-6}
Variant
& Epi.$\downarrow$
& Dur.$\downarrow$
& Start$\downarrow$
& Slot$\downarrow$
& CPC$\uparrow$ \\
\midrule
w/o Active-Day Gate ($A$)
& \ms{0.0029}{0.0006} & \ms{0.0016}{0.0004} & \ms{0.0047}{0.0004} & \ms{0.00017}{0.00005} & \ms{0.9706}{0.0038} \\
w/o Act.-Stru. Motifs ($z$)
& \ms{0.0223}{0.0048} & \ms{0.0051}{0.0006} & \ms{0.0089}{0.0008} & \ms{0.00213}{0.00052} & \ms{0.8910}{0.0098} \\
w/o Hie. Gen. ($\mathbf{x}\!\rightarrow\!Y$)
& \ms{0.0029}{0.0006} & \ms{0.0017}{0.0003} & \ms{0.0069}{0.0010} & \ms{0.00070}{0.00011} & \ms{0.9067}{0.0057} \\
\midrule
\rowcolor{gray!10}
\textbf{Act-HiMotif}
& \best{\ms{0.0014}{0.0007}}
& \best{\ms{0.0013}{0.0001}}
& \best{\ms{0.0045}{0.0003}}
& \best{\ms{0.00011}{0.00006}}
& \best{\ms{0.9749}{0.0027}} \\
\bottomrule
\end{tabular*}

\vspace{5pt}

\begin{tabular*}{\textwidth}{@{\extracolsep{\fill}}lccccc@{}}
\toprule
\multicolumn{6}{c}{\textbf{Spatial generalization}} \\
\cmidrule(lr){1-6}
Variant
& Epi.$\downarrow$
& Dur.$\downarrow$
& Start$\downarrow$
& Slot$\downarrow$
& CPC$\uparrow$ \\
\midrule
w/o Active-Day Gate ($A$)
& \ms{0.0025}{0.0008} & \ms{0.0045}{0.0009} & \ms{0.0125}{0.0010} & \ms{0.00028}{0.00007} & \ms{0.9568}{0.0043} \\
w/o Act.-Stru. Motifs ($z$)
& \ms{0.0211}{0.0028} & \ms{0.0079}{0.0013} & \ms{0.0175}{0.0015} & \ms{0.00199}{0.00060} & \ms{0.8929}{0.0078} \\
w/o Hie. Gen. ($\mathbf{x}\!\rightarrow\!Y$)
& \ms{0.0052}{0.0023}
& \best{\ms{0.0022}{0.0007}}
& \ms{0.0160}{0.0006}
& \ms{0.00081}{0.00027}
& \ms{0.9170}{0.0096} \\
\midrule
\rowcolor{gray!10}
\textbf{Act-HiMotif}
& \best{\ms{0.0021}{0.0009}}
& \ms{0.0042}{0.0003}
& \best{\ms{0.0120}{0.0010}}
& \best{\ms{0.00015}{0.00009}}
& \best{\ms{0.9653}{0.0038}} \\
\bottomrule
\end{tabular*}

\end{threeparttable}
\end{table*}

\subsection{Learned Activity-Structure Motifs}
\label{app:learned_motifs}

We analyze the motif codebook using active-day training schedules under both evaluation protocols. Coverage statistics are obtained by ranking motifs by training frequency within each run and aggregating the resulting ranks across five seeds. Concrete motif identities, dominant structures, and profile compositions are visualized using the fixed seed 2026 because motif code indices are not aligned across independently trained codebooks.

\begin{figure*}[!t]
    \centering
    \includegraphics[width=\textwidth]{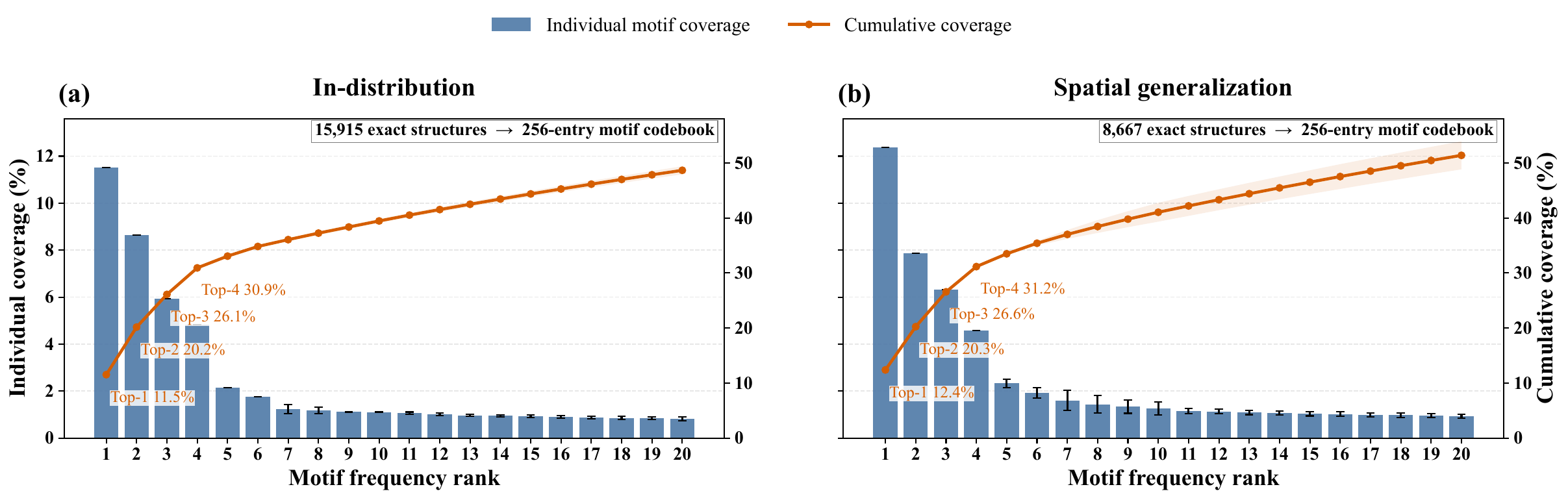}
    \caption{Frequency and cumulative coverage of the learned motifs. Bars report the mean individual coverage of the 20 most frequent motifs across five seeds, with error bars denoting one standard deviation. Curves and shaded regions show the corresponding mean cumulative coverage and across-seed variation. The in-distribution and spatial-generalization training sets contain 15,915 and 8,667 distinct exact activity structures, respectively, organized using a 256-entry motif codebook under each protocol.}
    \label{fig:motif_structure_space_coverage}
\end{figure*}

Figure~\ref{fig:motif_structure_space_coverage} shows a pronounced head--tail usage distribution. The four most frequent motifs account for approximately 31\% of active-day schedules under both protocols, while the top 20 motifs cover approximately half. Thus, the 256-entry codebook organizes thousands of distinct exact activity structures while retaining a broad tail of lower-frequency motifs.

\begin{figure*}[!t]
    \centering
    \includegraphics[width=\textwidth]{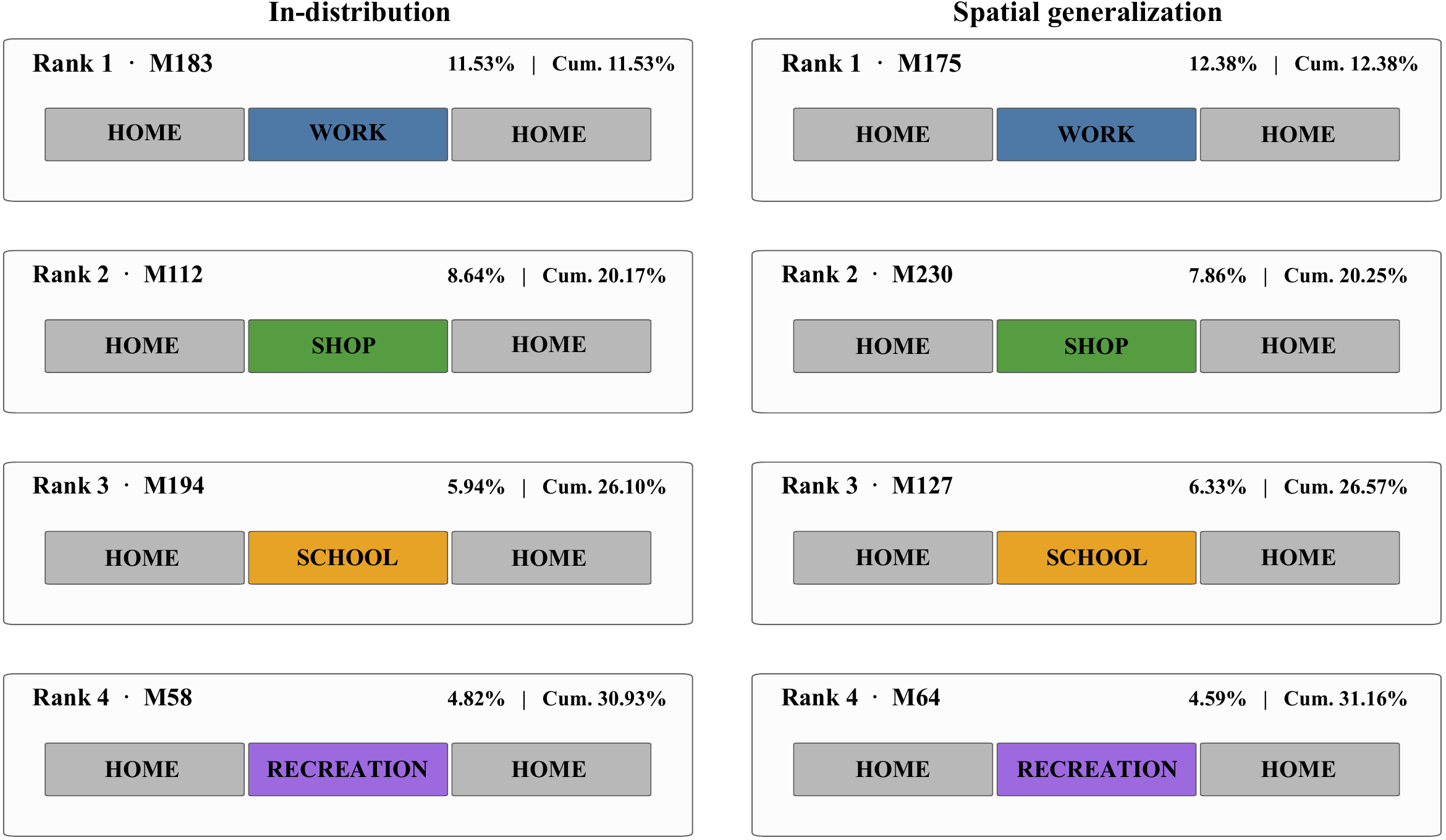}
    \caption{Dominant activity structures of the four most frequent motifs for the fixed canonical seed 2026. Each row reports the motif-frequency rank, run-specific code index, individual and cumulative training coverage, and dominant exact activity structure.}
    \label{fig:top4_motif_structures}
\end{figure*}

As shown in Figure~\ref{fig:top4_motif_structures}, the four most frequent motifs have the same dominant activity structures under both protocols: \texttt{H-W-H}, \texttt{H-SHOP-H}, \texttt{H-S-H}, and \texttt{H-REC-H}. Together, these motifs cover 30.93\% and 31.16\% of active-day training schedules under the in-distribution and spatial-generalization protocols, respectively. Although their code indices differ across independently trained codebooks, the recurring work-, shopping-, school-, and recreation-centered structures are consistent across protocols.

\begin{figure*}[!t]
    \centering
    \includegraphics[width=\textwidth]{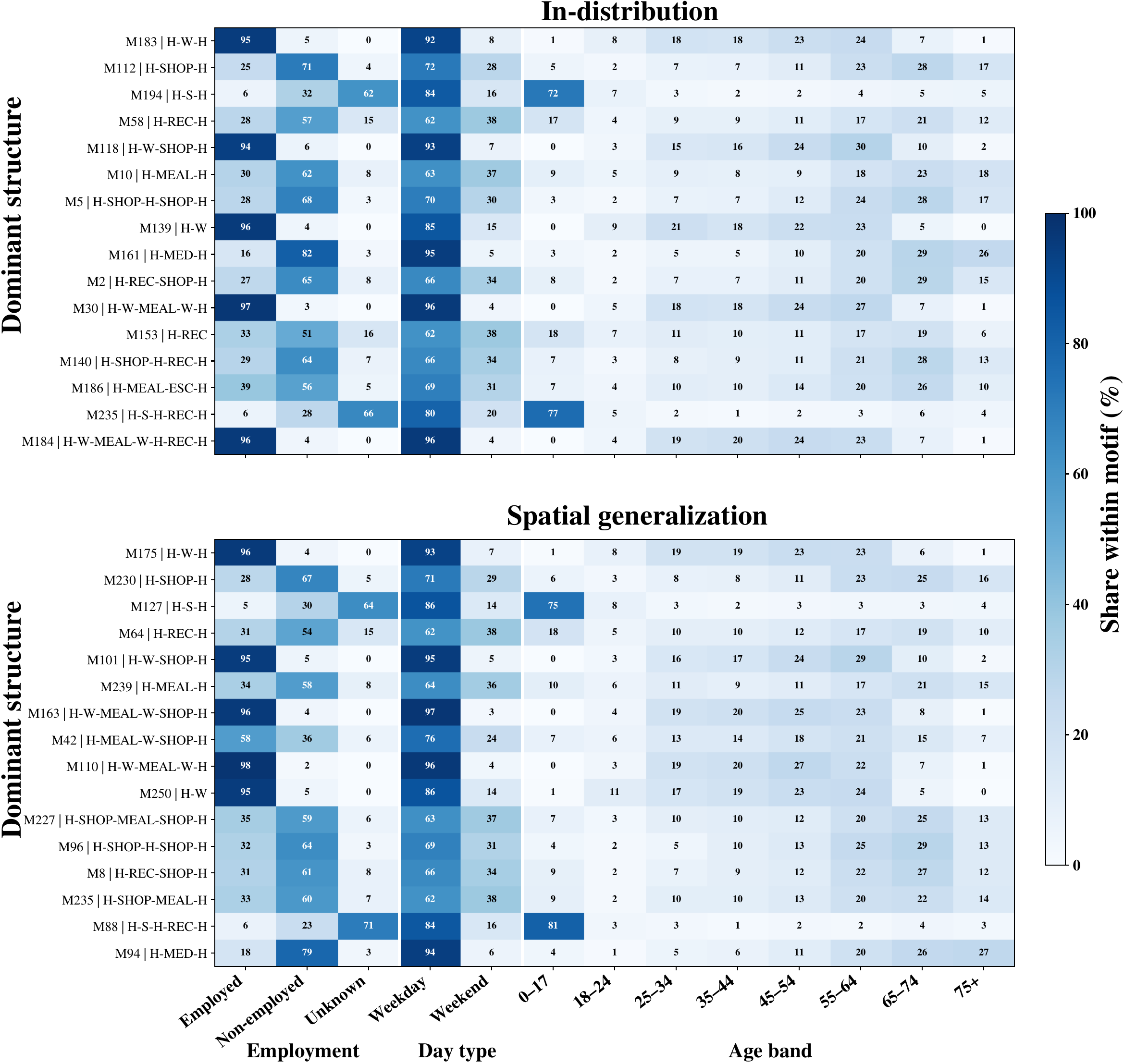}
    \caption{Profile composition of the 16 most frequent motifs (frequency ranks 1--16) for the fixed canonical seed 2026 under each evaluation protocol. Rows are ordered by active-day training frequency and labeled with the motif index and dominant exact structure. For each motif, percentages sum to 100\% separately across employment categories, day types, and age bands.}
    \label{fig:motif_profile_distribution}
\end{figure*}

Figure~\ref{fig:motif_profile_distribution} reports the profile compositions of the 16 most frequent motifs, ordered by active-day training frequency for the fixed seed 2026. The top-16 motifs exhibit distinct profile compositions under both evaluation protocols. Work-centered motifs such as \texttt{H-W-H} and \texttt{H-W-SHOP-H} contain large shares of employed individuals and weekdays, whereas the school-centered \texttt{H-S-H} motif is concentrated in the 0--17 age group. Shopping- and recreation-centered motifs contain larger shares of non-employed individuals or weekend schedules, while medical-centered motifs are more prevalent among older age groups. These qualitative associations are broadly consistent between the two evaluation protocols.

Overall, Figures~\ref{fig:motif_structure_space_coverage}--\ref{fig:motif_profile_distribution} show that the learned codebook provides a compact and interpretable organization of the long-tailed exact-structure space, with frequent motifs corresponding to recurring daily structures and distinct profile compositions.

\section{D. Profile-Conditioned Analyses}
\label{app:profile_analysis}

\subsection{Profile--Activity-Chain Alignment}

For each profile contrast, the prevalence difference is computed in percentage points among active-day schedules. Positive values indicate higher prevalence in the group shown on the right side of each panel, and negative values indicate higher prevalence in the group shown on the left. The four largest positive and four largest negative training-set differences are displayed.

\begin{figure*}[!t]
    \centering
    \includegraphics[width=\textwidth]{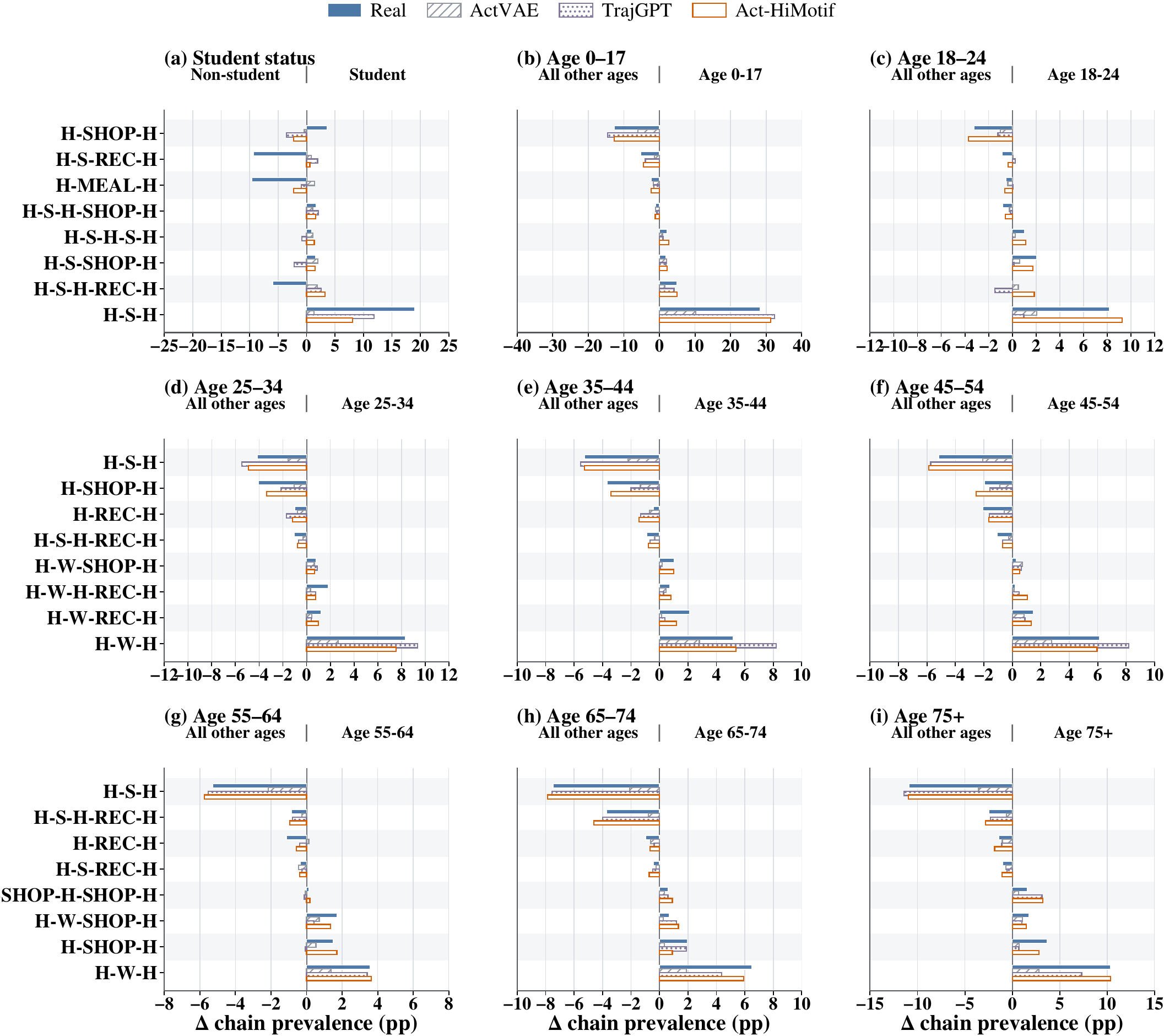}
    \caption{Profile--activity-chain alignment for student-status and age-band contrasts under CBSA-disjoint spatial generalization. Displayed active-day chains are selected using training-set prevalence differences.}
    \label{fig:app_chain_alignment_cbsa}
\end{figure*}

\begin{figure*}[!t]
    \centering
    \includegraphics[width=\textwidth]{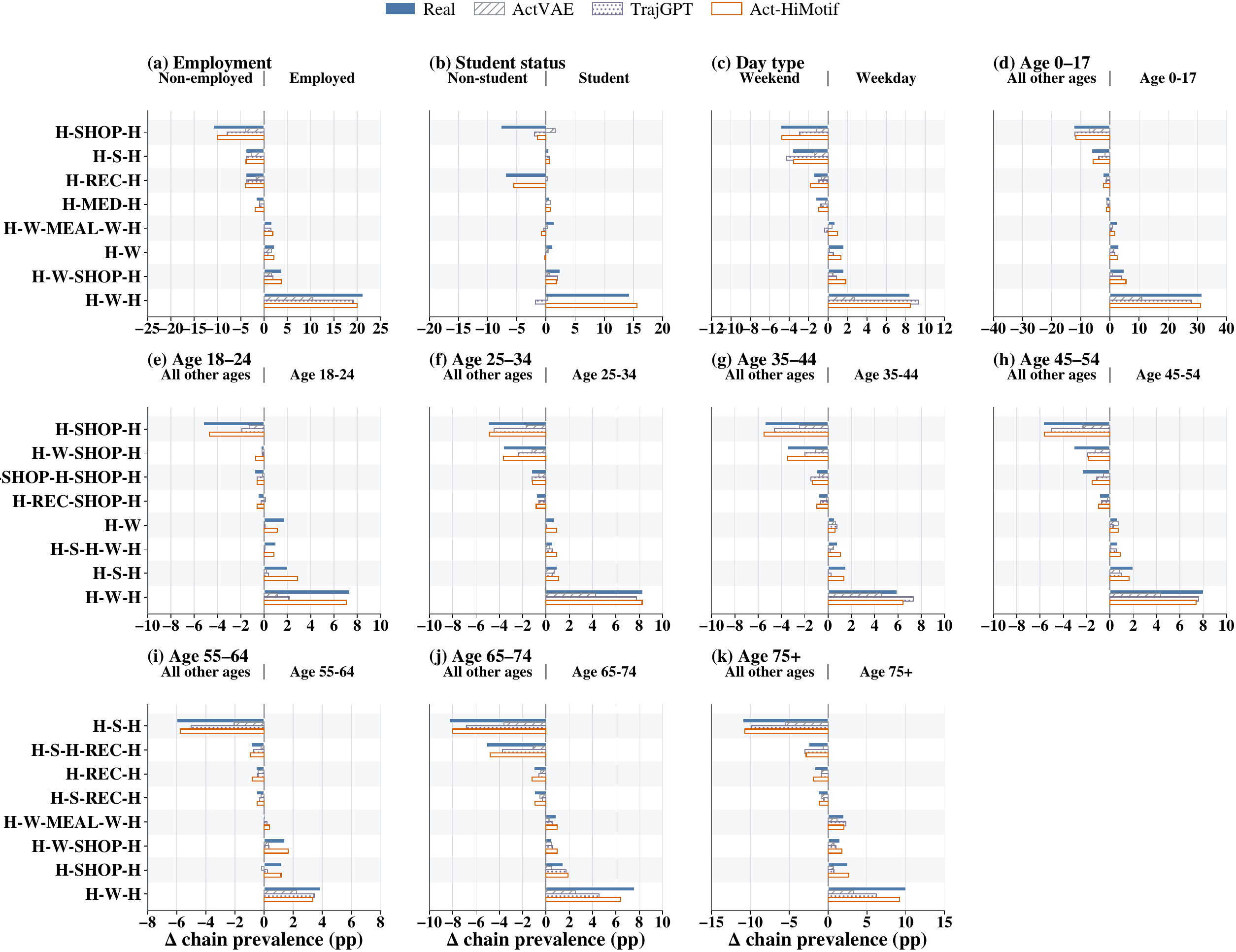}
    \caption{Profile--activity-chain alignment for all analyzed profile contrasts under in-distribution evaluation. Displayed active-day chains are selected using training-set prevalence differences.}
    \label{fig:app_chain_alignment_random}
\end{figure*}

Figure~\ref{fig:app_chain_alignment_cbsa} reports student-status and age-band contrasts under CBSA-disjoint spatial generalization. School-centered chains, particularly \texttt{H-S-H}, show strong positive prevalence differences for students and the 0--17 age group, while several older age bands show negative contrasts for these chains. Figure~\ref{fig:app_chain_alignment_random} shows corresponding patterns under in-distribution evaluation, including the strong association of \texttt{H-W-H} with employed individuals and \texttt{H-S-H} with students and younger individuals. Across both figures, Act-HiMotif generally tracks the direction and relative magnitude of the observed chain-prevalence differences.

\subsection{Profile-Stratified Schedule Fidelity}

\begin{figure*}[!t]
    \centering
    \includegraphics[width=\textwidth]{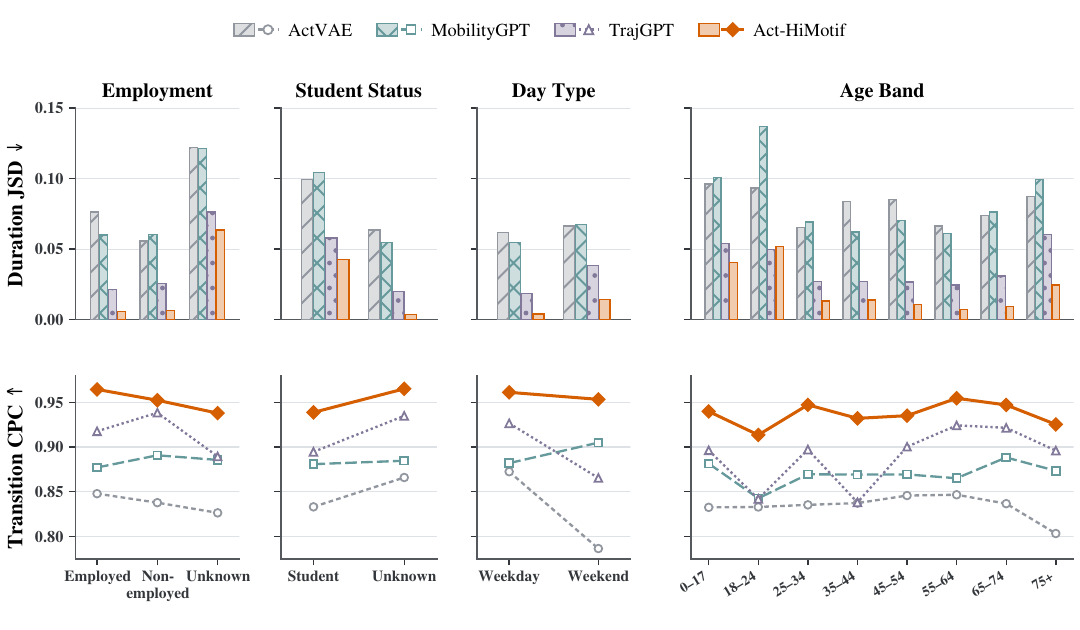}
    \caption{Profile-stratified Duration JSD and Transition CPC under CBSA-disjoint spatial generalization.}
    \label{fig:app_profile_stratified_cbsa}
\end{figure*}

\begin{figure*}[!t]
    \centering
    \includegraphics[width=\textwidth]{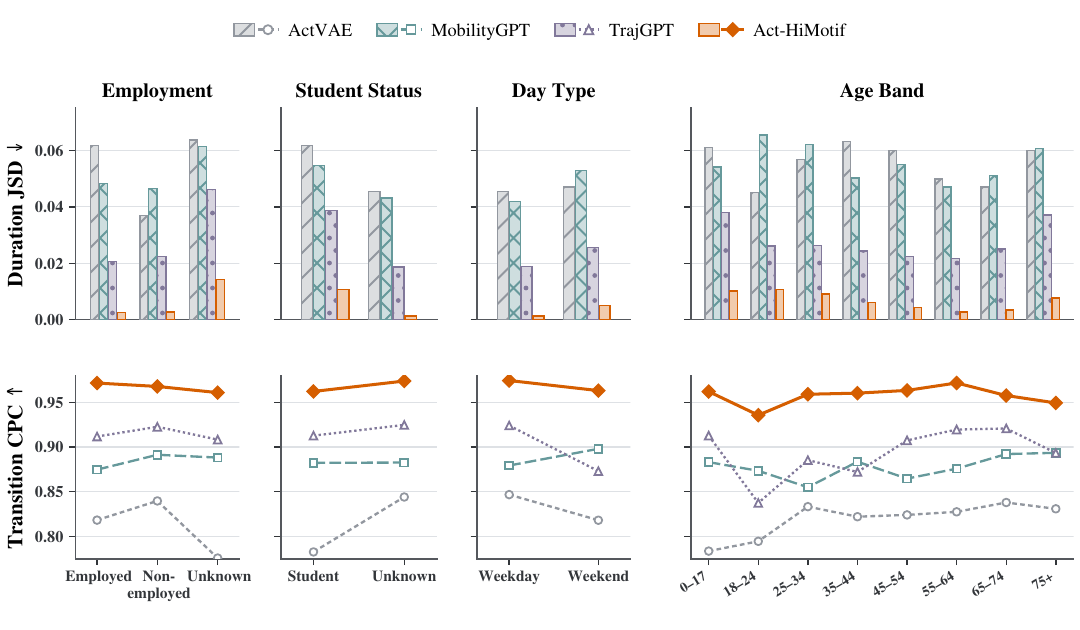}
    \caption{Profile-stratified Duration JSD and Transition CPC under in-distribution evaluation.}
    \label{fig:app_profile_stratified_random}
\end{figure*}

Figure~\ref{fig:app_profile_stratified_cbsa} reports subgroup-level results under CBSA-disjoint spatial generalization, and Figure~\ref{fig:app_profile_stratified_random} reports the corresponding in-distribution results. Across all retained employment, student-status, day-type, and age-band subgroups, Act-HiMotif achieves the lowest Duration JSD and highest Transition CPC under both protocols.

To avoid unstable subgroup-level distribution estimates, only groups with at least 500 test person-days are included in Figures~\ref{fig:app_profile_stratified_cbsa} and \ref{fig:app_profile_stratified_random}. The Non-student subgroup, with only 54 and 12 test person-days under the in-distribution and spatial-generalization protocols, respectively, is therefore omitted. This sample-size filter is applied identically to all methods and only to these subgroup-level distributional metrics.

\section{E. Structural--Temporal Analyses}
\label{app:structural_temporal}

\subsection{Joint Start-Time--Duration Distributions}

\begin{figure*}[!t]
    \centering
    \includegraphics[width=\textwidth]{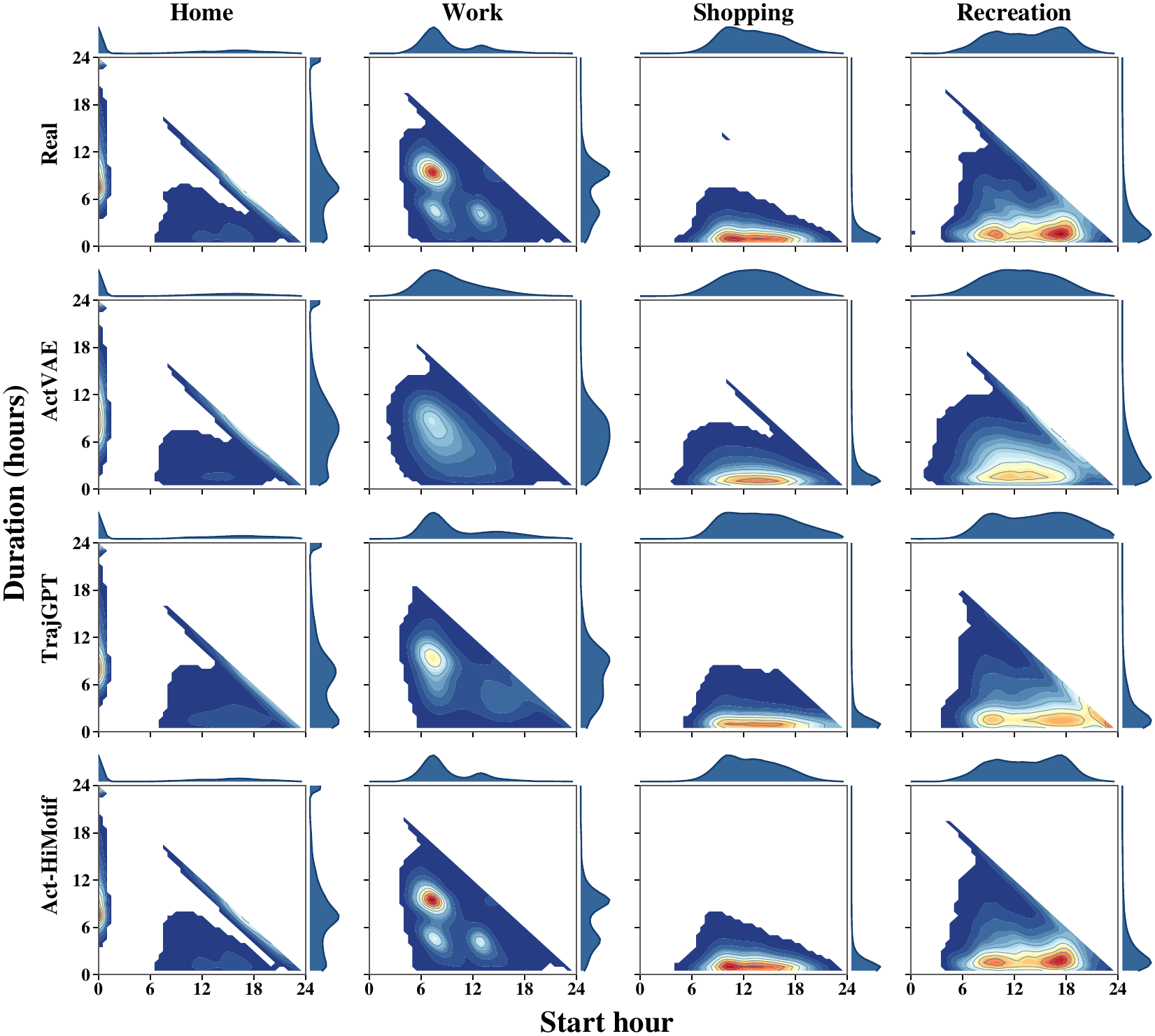}
    \caption{Observed and generated joint start-time--duration distributions for Home, Work, Shopping, and Recreation under in-distribution evaluation.}
    \label{fig:app_start_duration_random_four}
\end{figure*}

Figure~\ref{fig:app_start_duration_random_four} reports the joint start-time--duration distributions of Home, Work, Shopping, and Recreation under in-distribution evaluation. Home exhibits broad temporal support, Work contains concentrated morning-start modes with relatively long durations, Shopping is concentrated at shorter durations, and Recreation spans broader afternoon and evening periods. Act-HiMotif generally preserves these activity-specific shapes, including the multiple Work modes and the broader temporal structure of Recreation.

\begin{figure*}[!t]
    \centering
    \includegraphics[width=\textwidth]{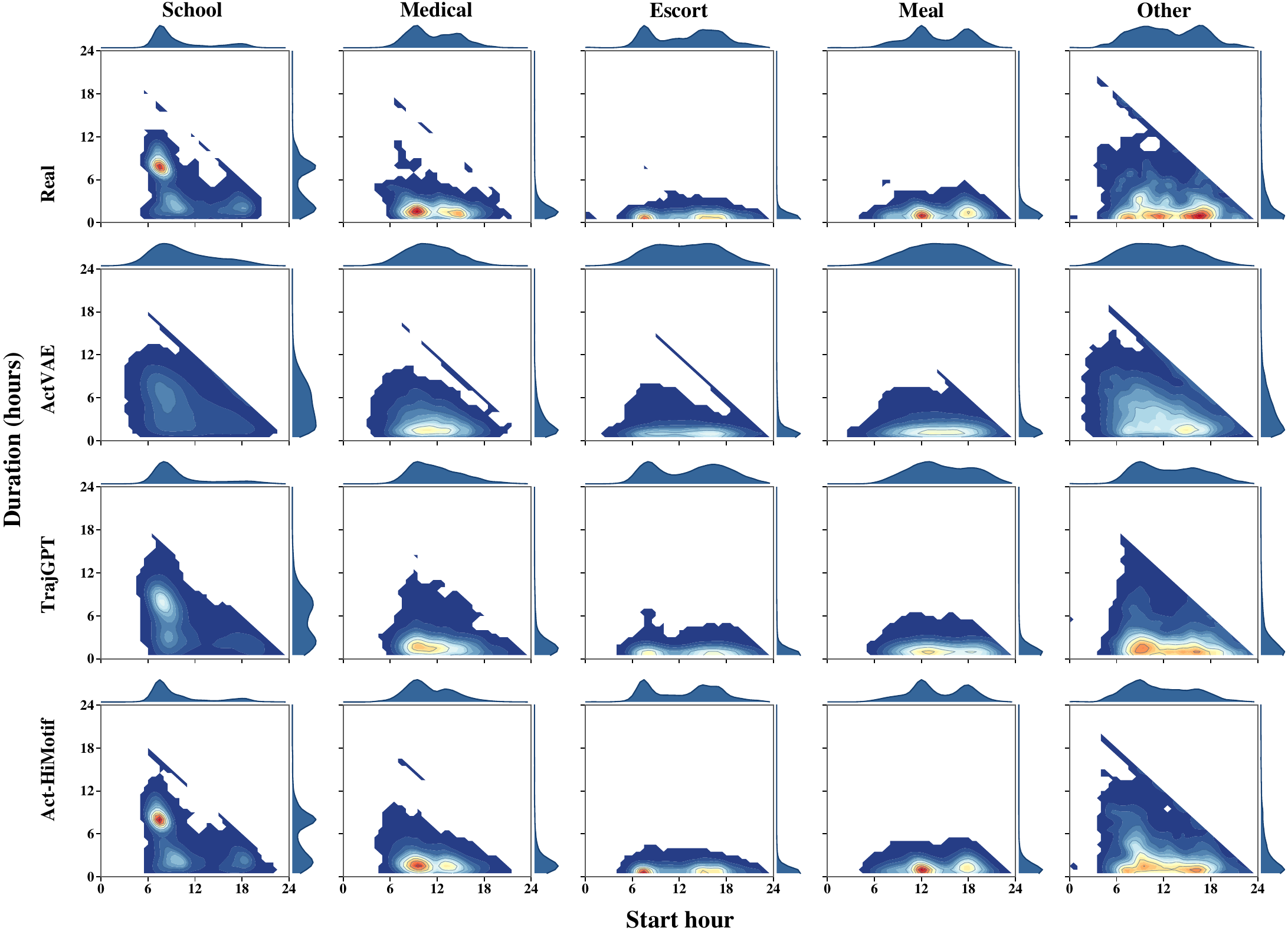}
    \caption{Observed and generated joint start-time--duration distributions for School, Medical, Escort, Meal, and Other under CBSA-disjoint spatial generalization.}
    \label{fig:app_start_duration_cbsa_remaining}
\end{figure*}

\begin{figure*}[!t]
    \centering
    \includegraphics[width=\textwidth]{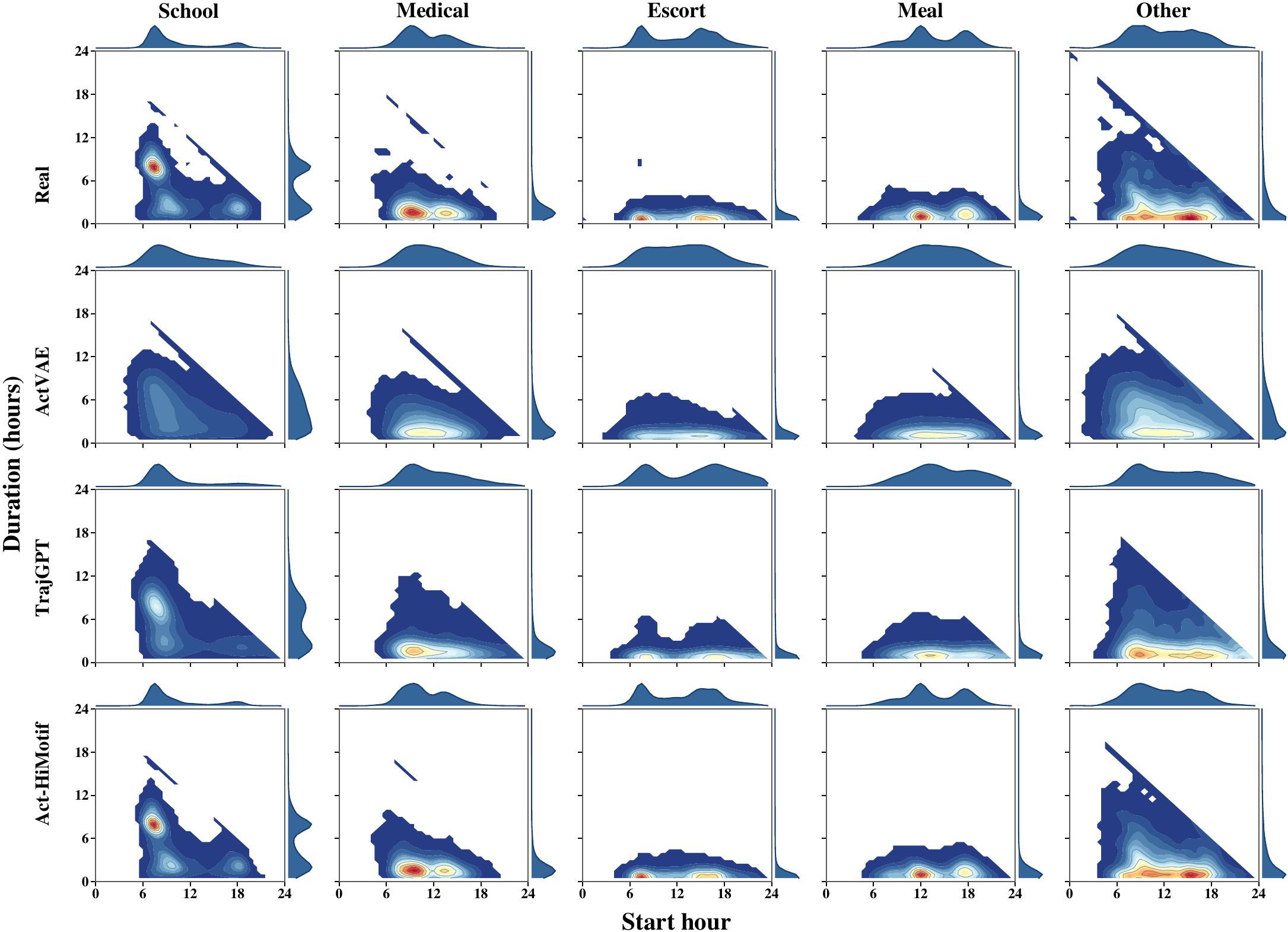}
    \caption{Observed and generated joint start-time--duration distributions for School, Medical, Escort, Meal, and Other under in-distribution evaluation.}
    \label{fig:app_start_duration_random_remaining}
\end{figure*}

Figures~\ref{fig:app_start_duration_cbsa_remaining} and \ref{fig:app_start_duration_random_remaining} report School, Medical, Escort, Meal, and Other under CBSA-disjoint spatial generalization and in-distribution evaluation, respectively. School exhibits a concentrated morning-start and relatively long-duration pattern, while Medical, Escort, and Meal are concentrated at shorter durations over daytime periods. Other spans a broader range of start times and durations. Across both figures, Act-HiMotif generally reproduces the locations and dispersions of these activity-specific patterns more closely, including the concentrated School pattern, the short-duration Escort
and Meal patterns, and the broader temporal support of Other.

\subsection{Activity-Transition Distributions}

\begin{figure*}[!t]
    \centering
    \includegraphics[width=\textwidth]{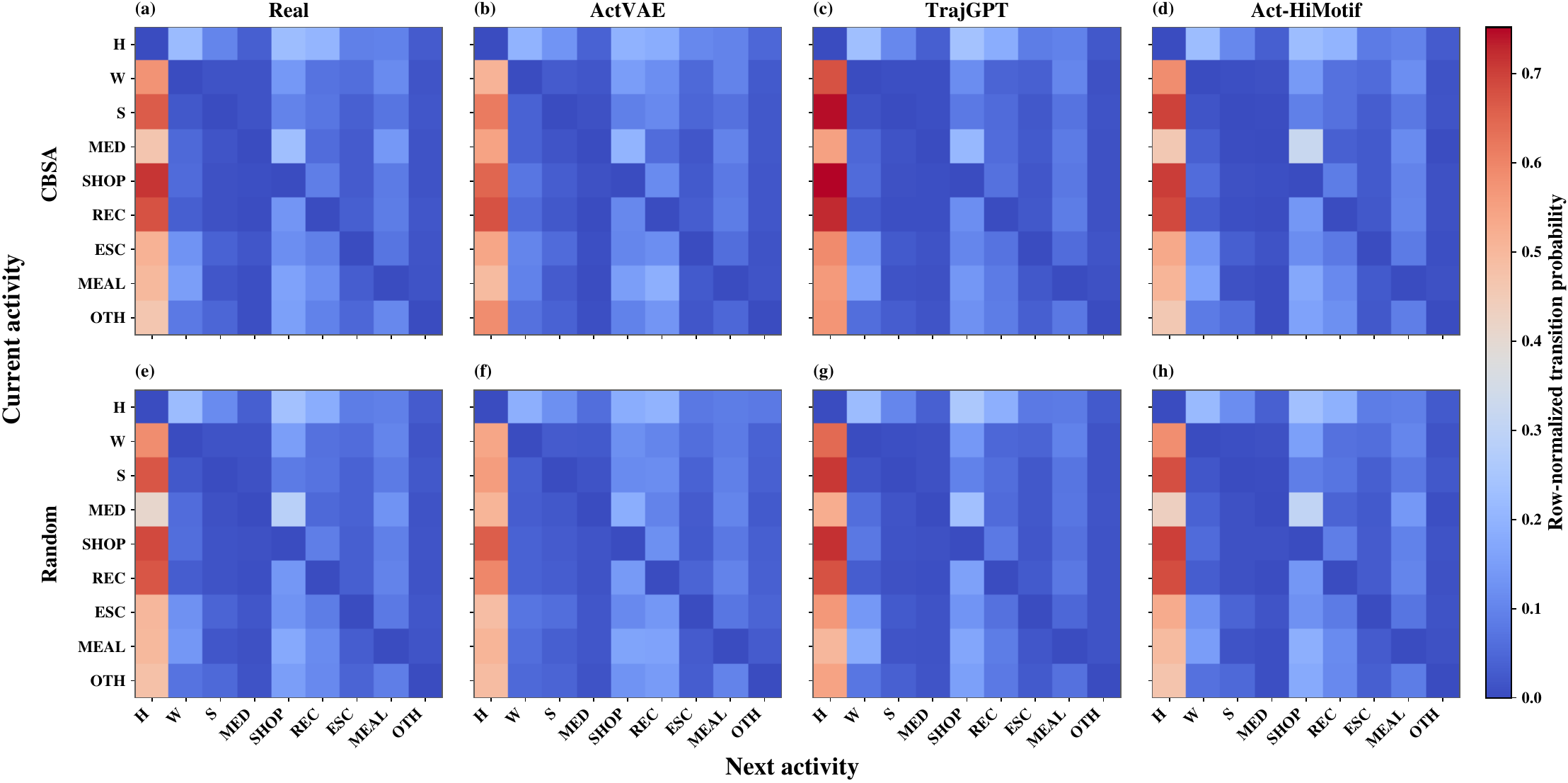}
    \caption{Row-normalized activity-transition distributions under CBSA-disjoint spatial generalization (top) and in-distribution evaluation (bottom). Rows denote current activities and columns denote next activities.}
    \label{fig:app_transition_matrices}
\end{figure*}

Figure~\ref{fig:app_transition_matrices} compares the row-normalized activity-transition distributions under both evaluation protocols, where each row represents the conditional distribution of the next activity given the current activity. The observed schedules contain strong Home-directed transitions together with activity-specific patterns among non-Home activities. ActVAE produces more diffuse transition patterns, while TrajGPT assigns more probability mass to a smaller set of Home-directed transitions than observed. Act-HiMotif more closely reproduces the dominant Home-related transitions and the relative organization of non-Home transitions under both protocols, consistent with its Transition CPC results.

\end{document}